\documentclass[letterpaper, conference]{IEEEtran}

\IEEEoverridecommandlockouts                              

\usepackage{graphics} 
\usepackage{epsfig} 
\usepackage{mathptmx} 
\usepackage{times} 
\usepackage{amsmath} 
\usepackage{amssymb}  
\usepackage{caption}
\usepackage{subcaption}
\usepackage{xcolor}
\usepackage{multirow}
\usepackage{booktabs}
\usepackage{adjustbox}
\usepackage{lipsum}

\DeclareGraphicsExtensions{.pdf, .PDF, .png, .PNG, .jpeg, .JPG}
\newcommand{\argmax}[0]{\arg\max}

\title{\LARGE \bf
Studying the Impact of Semi-Cooperative  Drivers   \\  on Overall Highway Flow}

\author{Noam Buckman$^{1}$, Sertac Karaman$^2$, Daniela Rus$^1$
\thanks{This work is supported by the Toyota Research Institute (TRI).  This article solely reflects the opinions and conclusions of its authors and not TRI or any other Toyota entity.  
The authors acknowledge the MIT SuperCloud and Lincoln Laboratory Supercomputing Center for providing HPC resources that have contributed to the research results reported within this paper.}
\thanks{$^{1}$Computer Science and Artificial Intelligence Laboratory,  Massachusetts Institute of Technology,  Cambridge,  MA 02139, USA 
{\tt\small [nbuckman, rus] at csail.mit.edu}}%
\thanks{$^{2}$Laboratory of Information and Decision Systems,  Massachusetts Institute of Technology,  Cambridge,  MA 02139,  USA
{\tt\small sertac@mit.edu}}%
}

\begin{document}

\newcommand{\work}{work}
\newcommand{\Conclusion}{Conclusion}

\maketitle
\thispagestyle{empty}
\pagestyle{empty}

\begin{abstract}

Semi-cooperative behaviors are intrinsic properties of human drivers and should be considered for autonomous driving.  
In addition,  new autonomous planners can consider the social value orientation (SVO) of human drivers to generate socially-compliant trajectories.
Yet the overall impact on traffic flow for this new class of planners remain to be understood. 
In this work, we present study of implicit semi-cooperative driving where agents deploy a game-theoretic version of iterative best response assuming knowledge of the SVOs of other agents.
We simulate nominal traffic flow and investigate whether the proportion of prosocial agents on the road impact individual or system-wide driving performance. 
Experiments show that the proportion of prosocial agents has a minor impact on overall traffic flow and that benefits of semi-cooperation  disproportionally affect egoistic and high-speed drivers.
\end{abstract}


\section{Introduction}
Improving traffic throughput with mixed human and autonomous vehicles has the potential to greatly reduce traffic congestion, reduce travel times, and improve driver experience. 
Many of these desirable improvements, however, are typically only achieved in scenarios where all vehicles are autonomous and controlled by a central planner. 
In the time being, the question remains whether we can gain traffic improvements in scenarios where the agents on the road do not explicitly communicate with each other but rather implicitly coordinate. 
In this \work{}, we consider deploying a semi-cooperative game-theoretic control algorithm on all driving agents so as to closely model the planning of each driving agent, both human and autonomous. 
By doing so, we can better understand the performance of such algorithms at a system-level and understand the impact of driver personality on the individual and system-wide performance.

In this \work{}, a system of semi-cooperative rational agents optimize their own individual reward function while considering the reward of other agents. 
Such a semi-cooperative model is observed in human participants for monetary games~\cite{liebrand1988SVO} and has been observed in highway driving~\cite{Schwarting2019}. 
However, in this \work{}  we consider the impact of the agent-specific cooperation or Social Value Orientation (SVO) on the overall system performance as we consider multiple planning agents on a highway setting. 
Specifically, we simulate highway driving under a variety of cooperative population, varying the number of cooperative (or prosocial) agents while measuring the the effect on road throughput as a function of cooperation level. 

Existing approaches to studying impacts of autonomous vehicles on overall traffic flow either consist of cooperation-agnostic driver models~\cite{Kesting2007a,Treiber2000} or fully cooperative fleets of vehicles~\cite{Lazar2019,Jin2018}, which do not capture the semi-cooperative nature of human drivers or autonomous vehicles. 
For example, autonomous vehicles may operate independently (not fleet operated) yet can consider semi-cooperative maneuvers. 
Other studies consider scenarios where individual autonomous vehicles optimize a system-wide utility function, which is typically not the case for individual traffic driving~\cite{Wu2017a,Toghi2022}. 
In contrast, this \work{} considers driving scenarios where individual drivers plan independently yet consider employing semi-cooperative controllers that consider the rewards of neighboring vehicles. 
Similar to human drivers, these controllers at times display cooperative maneuvers such as changing lanes to allow vehicles to pass while at times prioritizing their own reward function, maximizing vehicle speed.

In summary, we make the following contributions:

\begin{figure}
    \centering

    \begin{subfigure}{0.89\columnwidth}
        \centering
        \includegraphics[width=0.99\columnwidth]{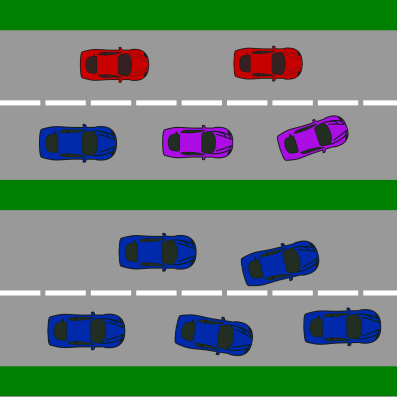}
    \end{subfigure}
    \caption[Varying Population by SVO]{Studying the overall traffic flow in systems with a mixture of prosocial and egoistic drivers (top) to systems with fully cooperative agents (bottom).}
    \label{fig:fig1}
\end{figure}

\begin{enumerate}
    \item A decentralized cooperative model predictive controller (MPC) for generating collision-free, semi-cooperative trajectories for human and autonomous drivers;
    \item Simulation of various cooperative populations of drivers with heterogeneous speed preferences deploying a semi-cooperative MPC;
    \item Study of the effect of the proportion of prosocial drivers on both system flow and individual driver speeds;
\end{enumerate}

\section{Related Works}

\subsection{Studying Traffic Flow in Mixed Human-AV Drivers}
Recent work has considered studying the traffic flow impacts of roadways with a mix of human and autonomous vehicles (AVs).
In \cite{Lazar2019}, researchers study the impact of autonomous drivers on the road to optimize traffic flow and influence human drivers.  
In that case, the planner is able to explicitly coordinate multiple AVs to optimize position in the traffic flow, whereas in this work we are interested in understanding both human and AV cooperation on vehicle performance.
For fully connected platooning vehicles,  the overall traffic capacity has been studied in~\cite{Jin2018} utilizing a fluid queuing model to study the interactions between vehicles at a traffic bottleneck. 
In~\cite{Wu2017a},  reinforcement learning agents optimize a system-level reward in a mixed human-autonomy traffic setting where a learned actions for a platoon of vehicles improves traffic throughput.
In~\cite{Wang2021}, stability and controllability of a mixed human-AV system is studied using control theory.
While emergent behaviors are studied,  human drivers are typically modeled as non-strategic or non-cooperative drivers using Intelligent Driver Model~\cite{Treiber2000} which model agents as vehicle-following agents. In contrast,  in this \work{}, we are interested in studying interactive planning which requires modeling the internal reward of each agent.  This allows us to better understand the impact of intrinsic human behavior on the overall speed of other vehicles.

\subsection{Non-Cooperative Driver Models}
The accurate modeling of human and autonomous agents is critical in better studying traffic flows,  especially in mixed human-AV environments.
One approach is to utilize microscopic simulators, such as SUMO~\cite{Lopez2018}, to study the performance of AVs and human drivers.
There,  human driver actions are modeled using vehicle following models such as Intelligent Driver Model (IDM)~\cite{Treiber2000} and MOBIL~\cite{Kesting2007a}. 
Much research have considered improved microscopic modeling of human drivers ~\cite{Calvert2020}, utilizing GANs to mimic human highway driving~\cite{Kuefler2017}, or closed-course field observations~\cite{Zhao2020}. 
While such models capture macroscopic traffic effects, they do not capture the underlying rational or optimization of the agents which is necessary when considering highly interactive maneuvers around other humans or autonomous vehicles. 
In addition, whereas some planners can consider the safety or risk of surrounding agents, they can not explicitly considering the rewards or reactions of non-ego vehicles~\cite{Glaser2010}.

\subsection{Fully Competitive Driver Models}
Alternatively,  humans and AVs can be been modeled as fully competitive agents that share a common environment with competing goals of traveling as fast as possible.
In such a framework, the AV models the internal reward of the other agents and deploy game-theoretic planners such as Iterative Best Response (IBR) or Reinforcement Learning (RL) to obtain agent actions.  
In~\cite{Fisac2019}, planning is split between high-level strategy and low-level,  short-horizon tactic for highway driving.
In~\cite{Wang2019}, the vehicles are fully competitive with a sensitivity term in the ego vehicle's reward function to capture interactions. 
In both,  drivers are considered to be fully competitive which is not necessarily the cooperative state of human drivers, given that humans typically show a range of cooperation levels.

\subsection{Semi-Cooperative Driver Models}
Semi-cooperative planning algorithms are a new class of autonomous planners that consider semi-cooperative reward structures of human drivers.
Schwarting et. al.~\cite{Schwarting2019} first proposed using social value orientation for modeling semi-cooperative behaviors in human drivers,  showing that driver actions in the NGSIM dataset can be jointly predicted with social value orientation. 
SVO has been extended to driving through intersections~\cite{Buckman2019},  interacting with pedestrians~\cite{Crosato2021},  ambulance driving~\cite{Buckman2021}, and merging with traffic~\cite{Toghi2022}.

Most similar to this work, ~\cite{Toghi2022} takes a multiagent reinforcement learning approach and focuses on training autonomous vehicles to cooperate according to the SVOs of AVs and human vehicles.  There, humans are modeled with IDM and AVs utilize the SVO model,  where an optimal SVO value is considered.
In contrast,  we take an explicit game-theoretic optimization approach without offline pretraining,  to highlight the impact of single-shot,  uncoordinated behaviors to better understand traffic flow. 
In addition,  we consider nominal highway driving instead of merging maneuvers and study the impact the the entire population's SVO makeup on the overall traffic movement. 

This \work{} follows up on the approach in ~\cite{Buckman2021} where a semi-cooperative MPC with Imagined Shared Control is first proposed.  
There,  an autonomous emergency vehicle utilized a modified version of iterative best response to navigate around human drivers. In contrast, this work considers a system of equally planning human drivers,  a more common setting for planning algorithms. In addition, we closely investigate the performance on the entire system and each agent in the system, not solely the a single autonomous emergency vehicle.

\begin{figure}[thb]
	\centering
	\begin{subfigure}{0.9\columnwidth}
	\centering
		\includegraphics[width=0.99\columnwidth, trim={200px, 75px, 225px, 125px}, clip]{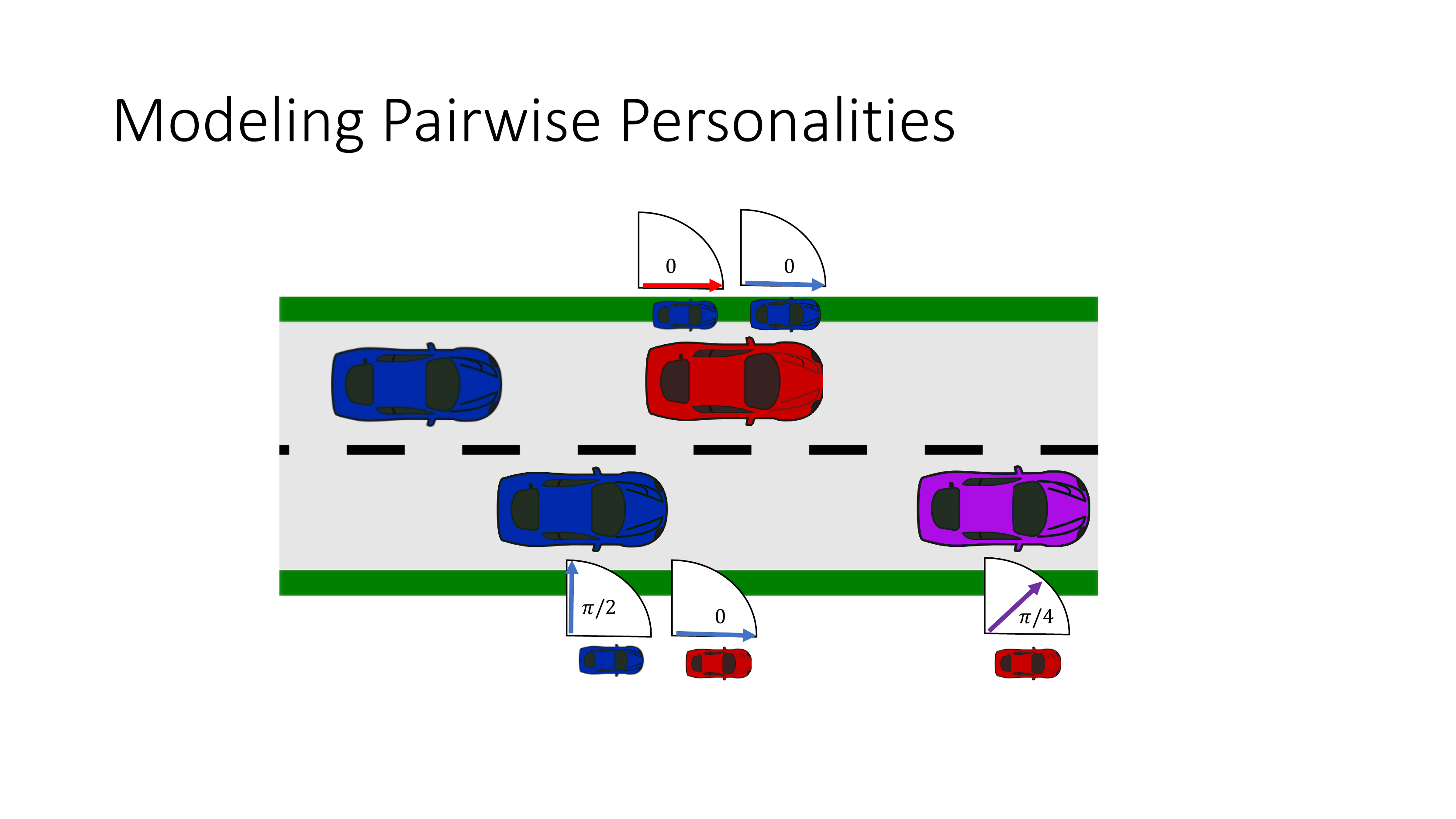}
		\caption{Pairwise Social Value Orientation}\label{fig:pairwise_svo}
	\end{subfigure}

	\begin{subfigure}{0.9\columnwidth}
	\centering
		\includegraphics[width=0.99\columnwidth, trim={200px, 75px, 225px, 150px}, clip]{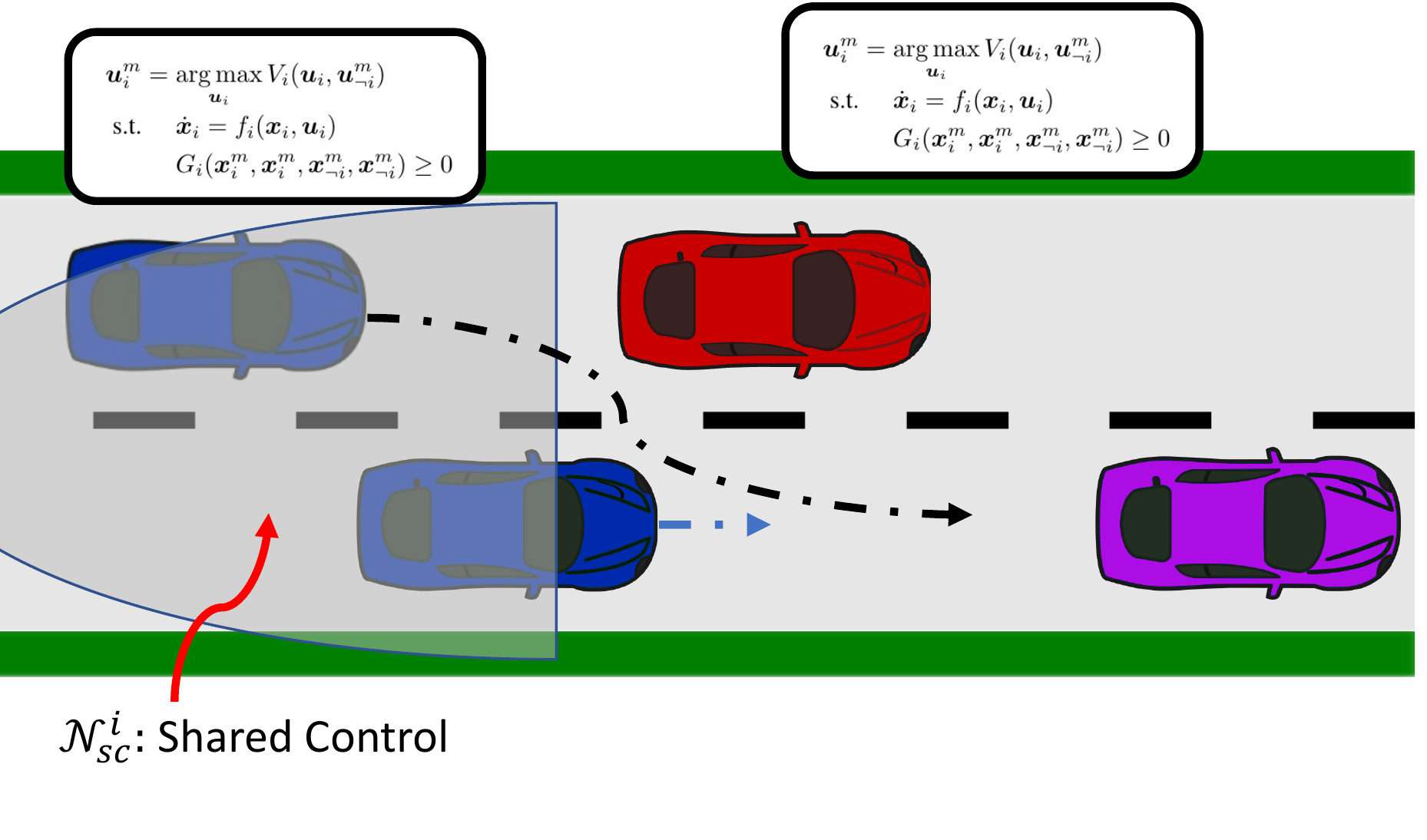}
		\caption{Iterative Best Response with Imagined Shared Control}\label{fig:shared_ibr}
	\end{subfigure}
	\caption[Semi-Cooperative Autonomous Planning]{Semi-Cooperative Autonomous Planning.  (\textbf{a})~Each agent generates a trajectory while optimizing a semi-cooperative optimization using pairwise social value orientation (SVO) between agents.  (\textbf{b})~Each agent converges to a Nash equilibrium control input using a modified version of iterative best response.}
    \label{fig:ibr1}	
\end{figure}

\section{Semi-Cooperative Planning for Humans}\label{sec:semi_cooper_ibr_iv}

\subsection{Problem Statement}
A system of agents $i=1\ldots n_{agents}$ must each independently generate control inputs $u_i$ while maintaining collision-free trajectories $\xi_i$.  Each agent's utility function is semi-cooperative of the form 
\begin{equation}
    V_i = \sum_{j\neq i} \cos \theta_{ij} R_i(u_i, x_i, x_j) + \sin \theta_{ij}R_j(u_j, x_j, x_i)
\end{equation}where $R_i(u_i, x_i, x_j)$ is a driver performance reward function based on its own (ego vehicle) control effort $u_i$, state $x_i$ and non-ego (ado vehicle) state $x_j$, and $\theta_{ij}$ is the pairwise SVO between ego agent $i$ and ado agent $j$. 

Each agent's dynamics $\dot{x_i} = f_i(x_i, u_i)$ are modeled by a Kinematic Bicycle Model with control inputs $u_i = [\delta_u, v_u]$ steering change and velocity change. We assume that all the vehicle have different speed limits, corresponding to an inherent heterogeneity that induces the need for cooperation. 
Each agent is assumed to optimize its own social utility function
\begin{align}
    u_i^* = \argmax_{u_i}  &  \: V_i(u_i, u_{\neg i}, x_i, x_{\neg i}) \nonumber \\ 
     \text{s.t. }  & \dot{x}_i = f_i(x_i, u_i)  \nonumber \\
                   & E_i \notin \cap E_{\neg i} \nonumber \\
     & u_i \leq u_{max} \label{eq:sims_problem_optimization}
\end{align}
where $f_i$ are the vehicle dynamics, $E_i$, $E_{\neg i}$ are 2D birds-eye-view bounding ellipses circumscribing each vehicle, and $u_{\max}$ are the control effort limits.  

The vehicle-specific performance $R_i$, which appears both in the ego vehicle utility function $V_i$ and ado vehicle utility function $V_{\neg i}$, is a linear combination of costs corresponding to maximizing speed, trajectory tracking, avoiding collisions, and conserving control effort
\begin{align}
    R_{i} = k_v ||v_i||^2  - & k_u ||u_i||^2 - k_{speeding} l_{speeding}^2  \\
     & - k_{kat} e_{lat}^2 -
     k_{lon} e_{lon}^2  - k_{ttc} C_{ttc} \nonumber
\end{align}
where $e_{lat}$ and $e_{lon}$ are the lateral and longitudinal errors from a desired trajectory, $||u||$ is the L2-norm on control effort, and $l_{speeding}$ is a speed-slack variable that penalizes vehicles that surpass their personal speed limit, such that $l_{sp} = 0$ if $v_i \leq v_{\max}$ else $l_{speeding} = (v_{i} - v_{\max})^2$. $C_{ttc}$ is a time-to-collision cost that we will describe in detail, in Sec.~\ref{sec:safety}. 
The reward function $R_i$ is evaluated at each time step $t$ of the optimization, however, for simplicity, we exclude the subscript $t$.

One significant difference in our setup from ~\cite{Buckman2021} is that vehicles are heterogeneous in both their semi-cooperative personality $\theta_{ij}$ and their desired speed (or speed limit) $v_{i,max}$, whereas in~\cite{Buckman2021}, only the emergency vehicle operated at higher speeds. One reason for considering heterogeneous speed limits is to consider scenarios where vehicles may have slightly different reward functions or vehicle dynamics. 
In addition, allowing for heterogeneous speed limits creates a more difficult control problem, given that differences in speed lead to both more dangerous driving scenarios (fast cars driving behind slow cars) and opportunities for cooperation, such as slower vehicle moving out of the way to allow faster vehicles to pass.

\subsection{Iterative Best Response with Imagined Shared Control}
For clarity, we summarize the Iterative Best Response with Imagined Shared Control, first introduced in ~\cite{Buckman2021},  where the vanilla iterative best response (IBR) is modified to consider both semi-cooperative agents and implicit coordinating for an emergency vehicle. 
In this \work{}, we extend previous work by considering homogeneous agents and additional safety considerations to better understand the impact of SVO on nominal driving.

The optimization in Eq.~\eqref{eq:sims_problem_optimization} is typically difficult to solve due to the nonlinear dynamics, non-stationary characteristic of $u_{\neg i}, x_{\neg i}$ or overall difficulty ensuring Nash equilibrium.  
A popular approach is to fix $u_{\neg i} = \bar{u}$ and solve the simplified problem 
\begin{align}
    u_i^* = \argmax_{u_i}  &  \: V_i(u_i, \bar{u}_{\neg i}, x_i, \bar{x}_{\neg i}) \label{eq:iv_br_opt} \\ 
     \text{s.t. }  & g(x) \geq 0  \\nonumber \label{eq:iv_br_constraints}
\end{align}
where $g(x) \geq 0$ captures both the equality constraint (dynamics) and inequality constraints (control limits, collision avoidance). 
This lowers the complexity in the optimization, however, cooperative solutions are no longer possible without an explicit cooperative cost. 
One approach is to rely heavily on the collision cost which includes $x_i, x_j$, however, they will not consider the control $u_j$. In addition, this leads to only considering the effects on one's own collision avoidance but not the effects on the other agent.  Alternatively, one can locate a local Nash equilibrium by solving the complete problem.  
We propose a middle ground where during iterative best response, agent $i$ solves for the control for a neighborhood of agents $j \in \mathcal{N}_{sc}$ but fixes the controls of any other agents.  
As iterative best response proceeds, the neighborhood size decreases $|\mathcal{N}_{sc}| \rightarrow 0$ such that by the end of iterative best response, agents are only solving for their own control. 
Figure~\ref{fig:ibr1} shows the main components of the Iterative Best Response with Imagined Shared Control algorithm, where agents are first modeled using a pairwise SVO~(Fig.~\ref{fig:pairwise_svo}) and then a neighborhood of shared control is considered during iterative best response to converge to more cooperative Nash Equilibrium~(Fig.~\ref{fig:shared_ibr}).
Previous work~\cite{Buckman2021} has shown that Nash equilibrium can be achieved, however the quality of those solutions were not explored for vehicles with heterogeneous speed limits and their personality. In this paper, we empirically explore the impact of this algorithm on vehicle performance.

\section{Ensuring Safe and Feasible Trajectories}\label{sec:safety}
\subsection{Safety Beyond a Finite Horizon}
The lack of guaranteed safety beyond the planning horizon is a significant limitation of finite horizon optimization, especially in the presence of vehicles that can drive at different top speeds. 
For example, a vehicle may drive at a speed that is collision-free for during the time horizon $T$ but leads to a collision at $t=T+\epsilon$.  
We address this issue by including a time-to-collision cost, similar to a control barrier function, that penalizes final speeds.

We first parameterize each vehicle's geometry with $k=2$ circumscribing circles centered at $p^{k}_{i} = [x^k_i, y^k_i]^T$ and corresponding radius $r^k_{i}$ along the length of the vehicle. 
We compute a pairwise modified time-to-collision cost for each pair $(p_i, p_j)$ of circles between agent $i$ and $j$ to account for the radii, as
\begin{align}
     p_{ij} = p_i - p_j \\
     v_{ij} = v_i - v_j \\
     d_{ij} = ||p_{ij}|| - r_{i} - r_{j} \\
    \tilde{p}_{ij} = p_{ij} \frac{||p_{ij}||}{||p_{ij}|| - r_i - r_j} \\ 
    t_{ttc} = f_{ttc}(\tilde{p}_{ij}, v_{ij})
\end{align}
where $v_i, v_j$ are the respective vehicle velocities, $f_{ttc}$ is the definition of time-to-collision
\begin{equation}
    f_{ttc}(p_{ij}, v_{ij}) = \frac{p_{ij}^T p_{ij}}{p_{ij}^T v_{ij}} \label{eq:ttc}.
\end{equation}

\begin{figure}[htb]
    \centering
    \includegraphics[width=0.89\columnwidth, trim={0px, 0px, 0px, 75px}, clip]{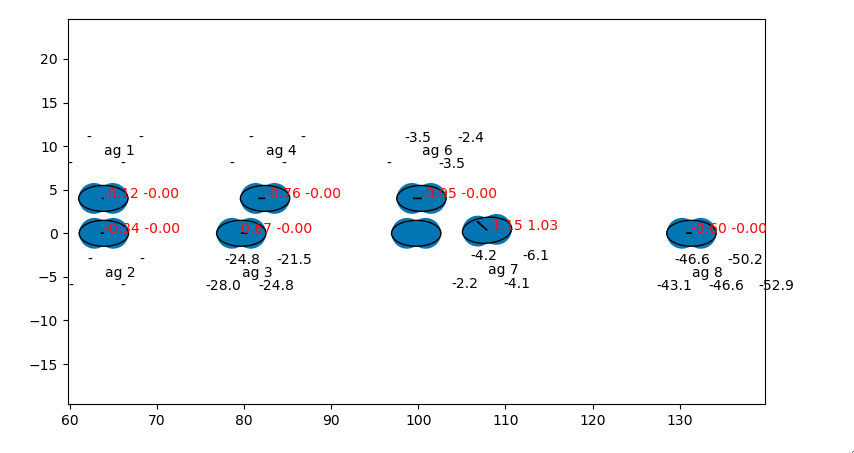}
    \caption[Time to Collision Cost]{Time to Collision Cost. Pairwise TTC cost between each circumscribing circle. Non-negative time to collision are designated with a '-'.}
    \label{fig:ttc_cost}
\end{figure}

In addition, we modify the time-to-collision cost in two ways, first by adding a velocity buffer and second, adding a scaling to the time-to-collision calculation to bias vehicles in the same lane as the ego vehicle, similar to the risk metric in~\cite{Pierson2018}. 
First we compute an indicator variable $F_{ij}$ of whether vehicle $j$ is in front of vehicle $i$ to determine whether to add a deceleration or acceleration buffer
\begin{align}\label{eq:velocity_buffer}
    d_{\phi} = [\cos(\phi), \sin(\phi)]^T \\
    F_i = \frac{\max ( - p_{ij}^T d_{\phi},0)}{-p_{ij}^T d_{\phi}} \\ 
    \tilde{v}_j = v_j \frac{ ||v_j|| + v_{\epsilon}(1 - 2 F_i)}{||v_j||} \\
    \tilde{v}_{ij} = v_i - \tilde{v}_j
\end{align}
where $d_{\phi}$ is the direction vector of the ego vehicle's orientation, $\tilde{v}_{ij}$ is the new relative velocity for computing the time-to-collision. 
Second, we bias the final time-to-collision metric in ~\eqref{eq:cosine_distance_scale} to more strongly penalize low time-to-collision with vehicles in the same lane compared to those driving in parallel lanes. 
We introduce a cosine-distance scaling to the time-to-collision 
\begin{equation}\label{eq:cosine_distance_scale}
    \tilde{t}_{collision} = \frac{f_{ttc}(\tilde{p}_{ij}, \tilde{v}_{ij})}{D_{cosine}(p_{ij}, d_\phi)}
\end{equation}
where $\phi$ is the orientation of the ego vehicle, $D_{cosine}(p_{ij}, d_{\phi}) = \frac{p_{ij}^T d_{\phi}}{||p_{ij}|| ||d_{\phi}||}$ is the cosine distance between the inter-vehicle distance vector and the ego vehicle direction vector. 
The final cost penalizes negative time-to-collisions 
\begin{equation}\label{eq:ttc_cost}
    C_{ttc} = \begin{cases}
        \frac{k_{ttc}}{\tilde{t}_{collision}^2} & \tilde{t}_{collision} < 0 \\ 
        0 & \tilde{t}_{collision} > 0 \\ 
    \end{cases}
\end{equation}
where $k_{ttc}$ is a time-to-collision cost weighting.
Figure~\ref{fig:ttc_cost} shows graphically the time to collision cost for circumscribing circles around cars in a traffic scenario. 
For each ado vehicle $j$ there are a total of 4 computations (for each pair of the two vehicle's circumscribing circles).

\begin{figure}[htb]
    \centering
    \includegraphics[width=0.99\columnwidth,trim={0 230px 0 245px},clip]{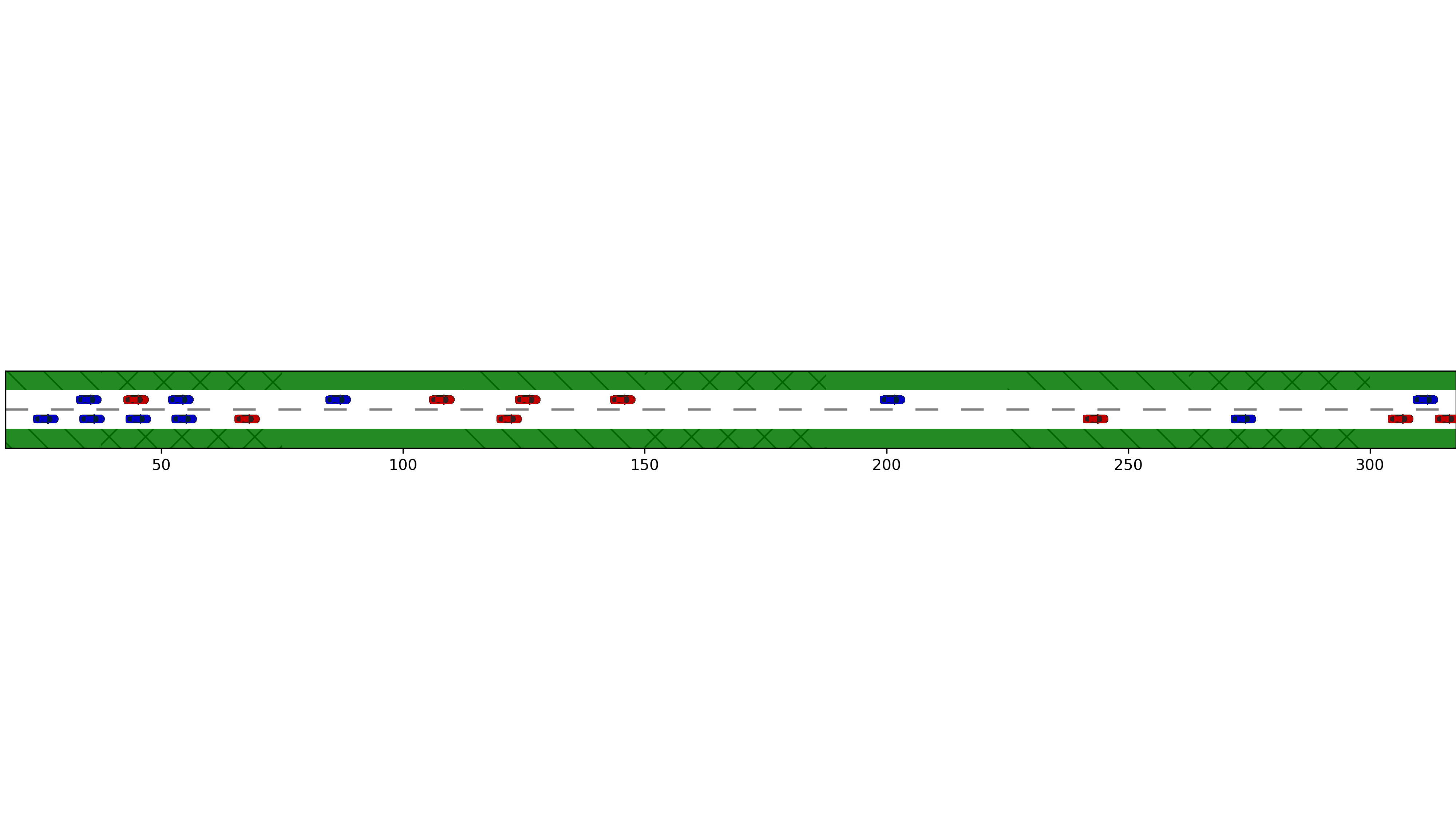}
    \includegraphics[width=0.99\columnwidth,trim={0 230px 0 245px},clip]{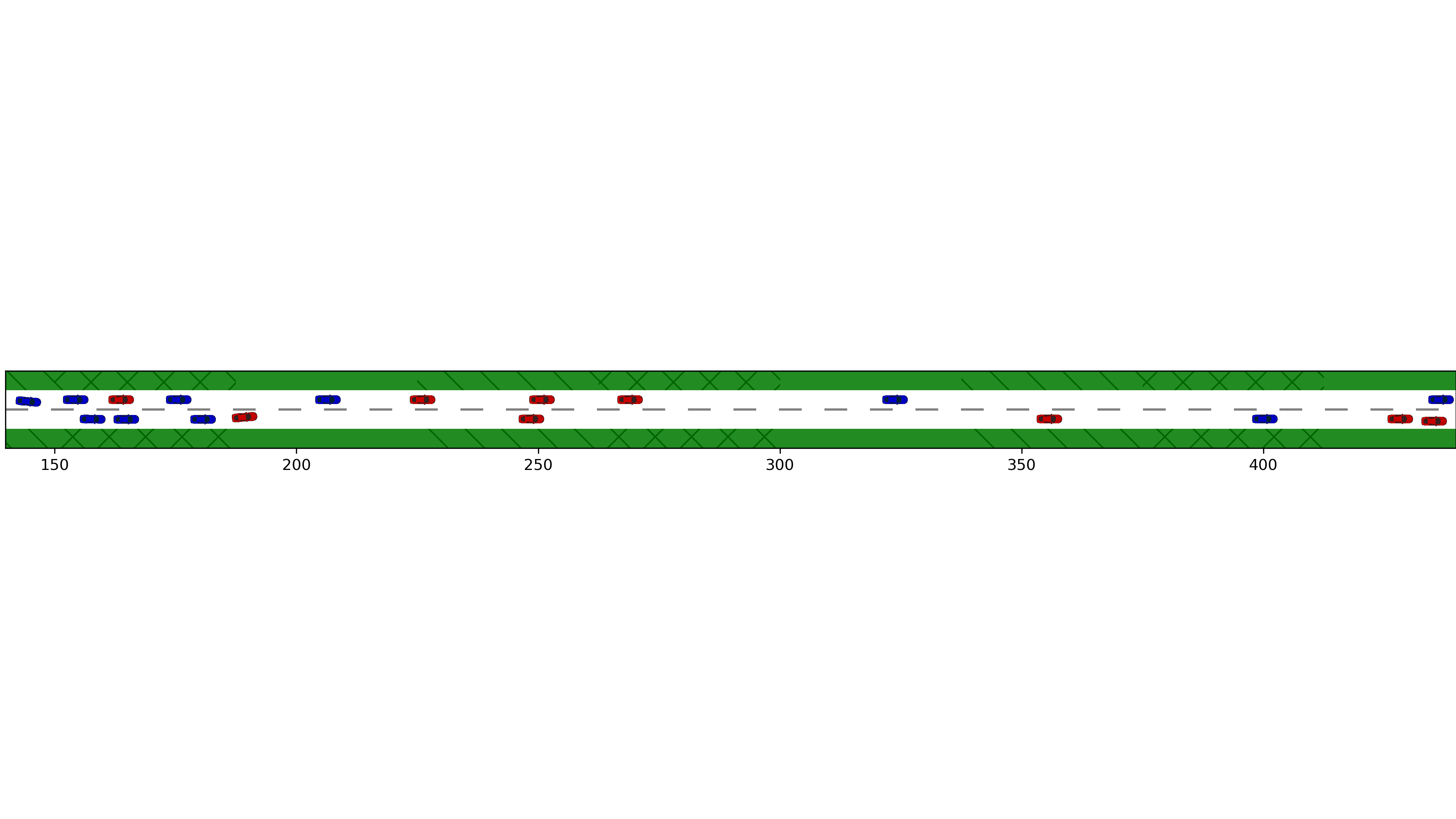}
    \includegraphics[width=0.99\columnwidth,trim={0 230px 0 245px},clip]{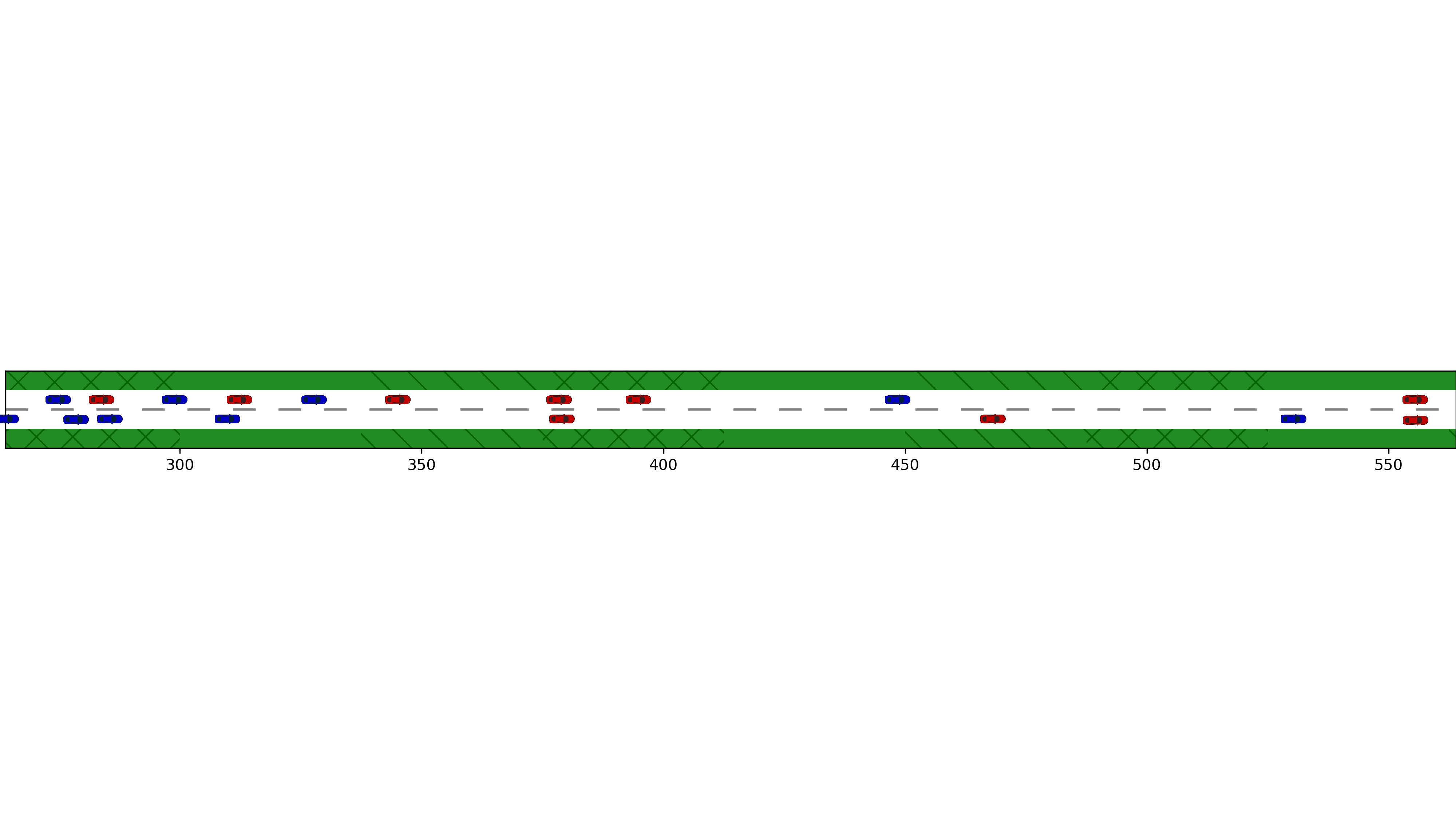}
    \includegraphics[width=0.99\columnwidth,trim={0 230px 0 245px},clip]{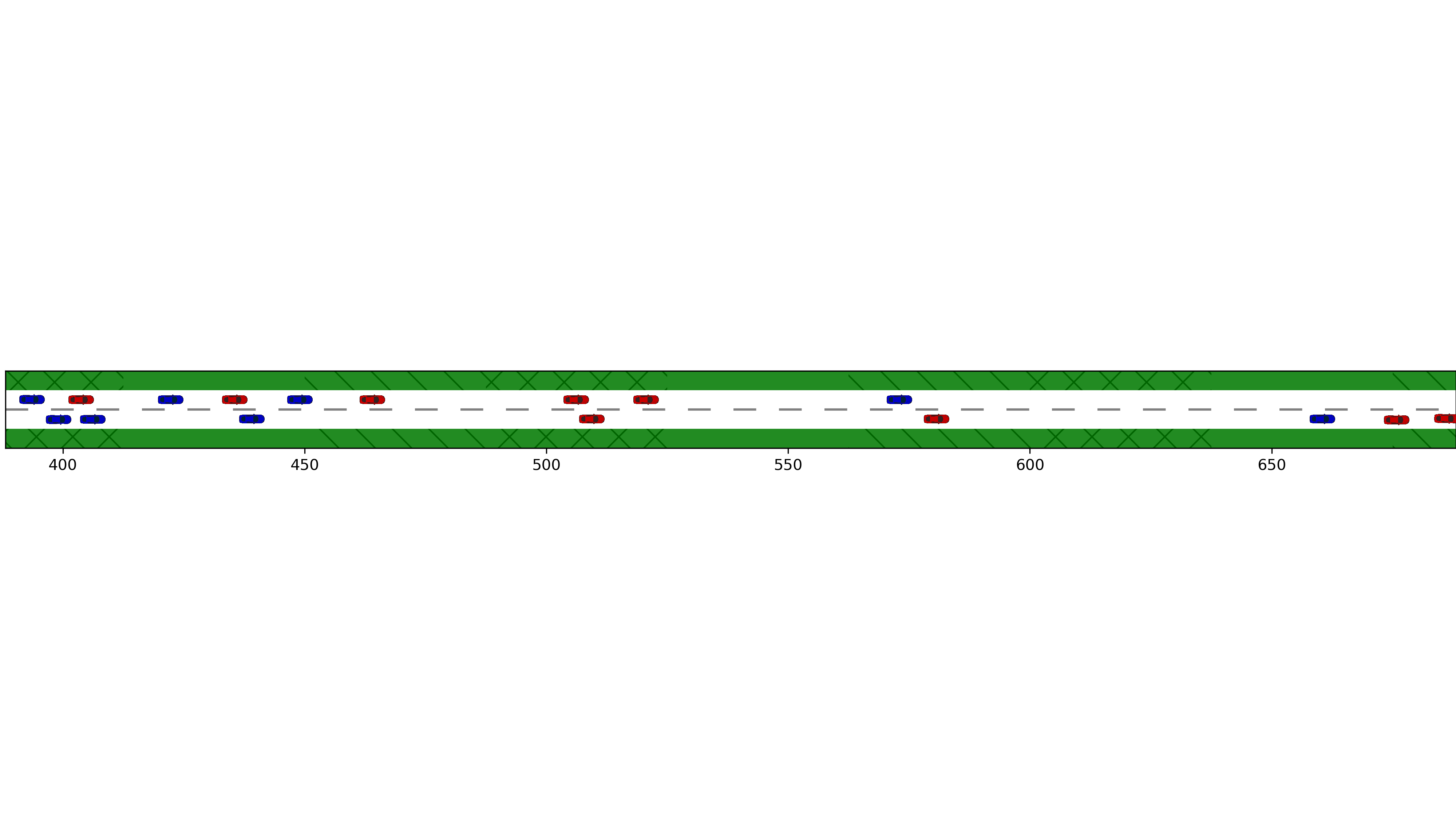}
    \includegraphics[width=0.99\columnwidth,trim={0 230px 0 245px},clip]{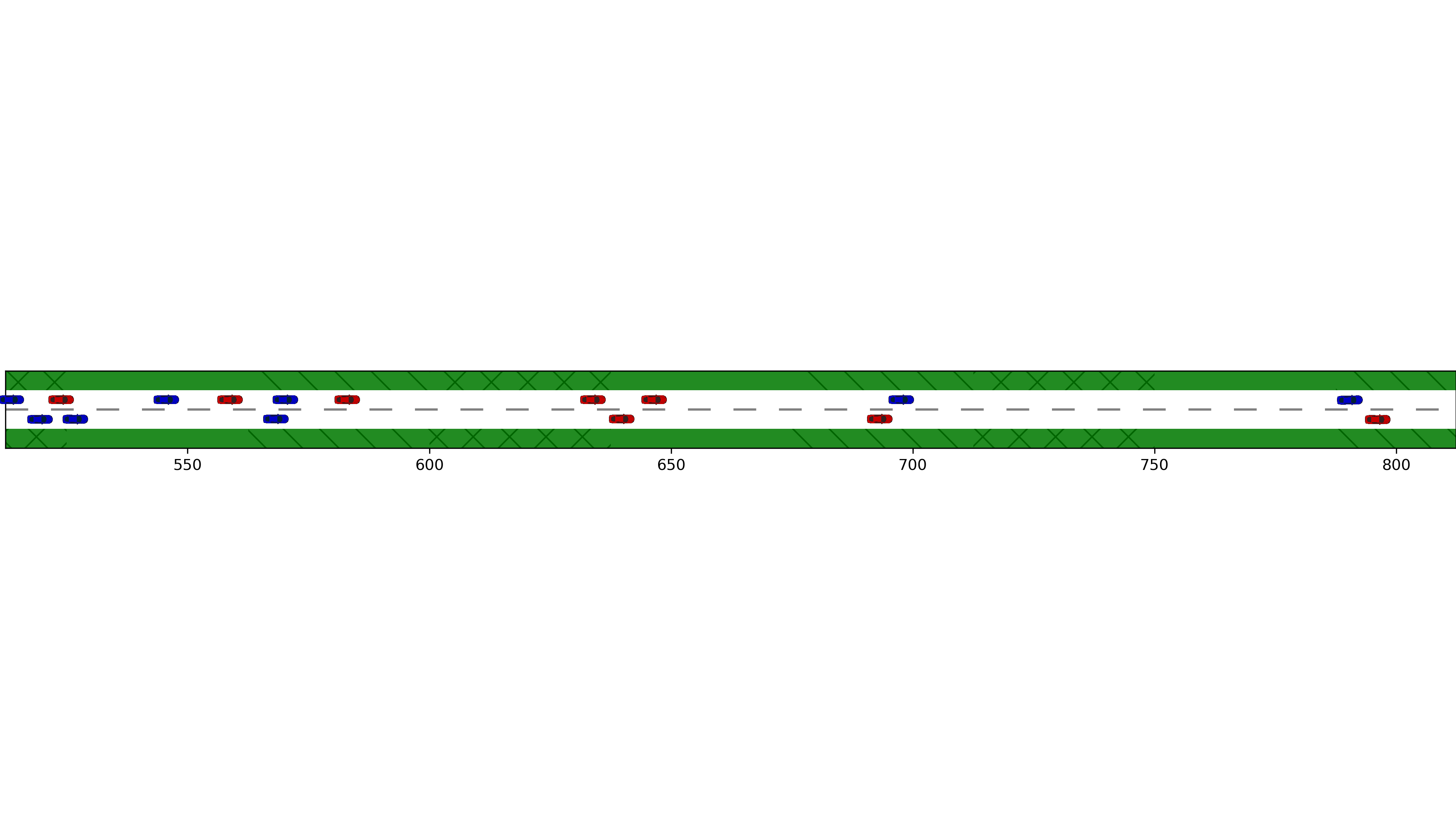}
    \includegraphics[width=0.99\columnwidth,trim={0 230px 0 245px},clip]{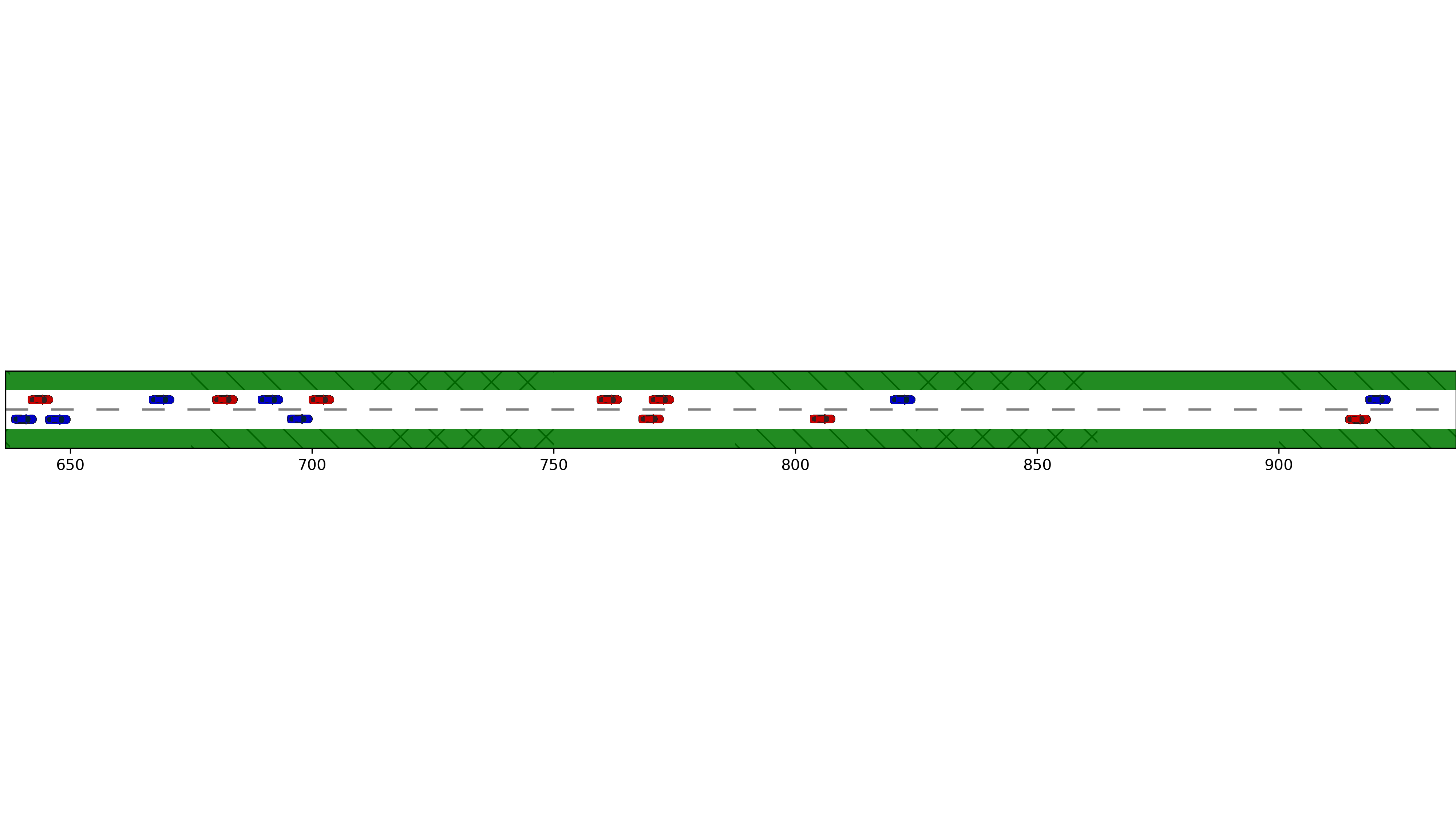}
    \includegraphics[width=0.99\columnwidth,trim={0 230px 0 245px},clip]{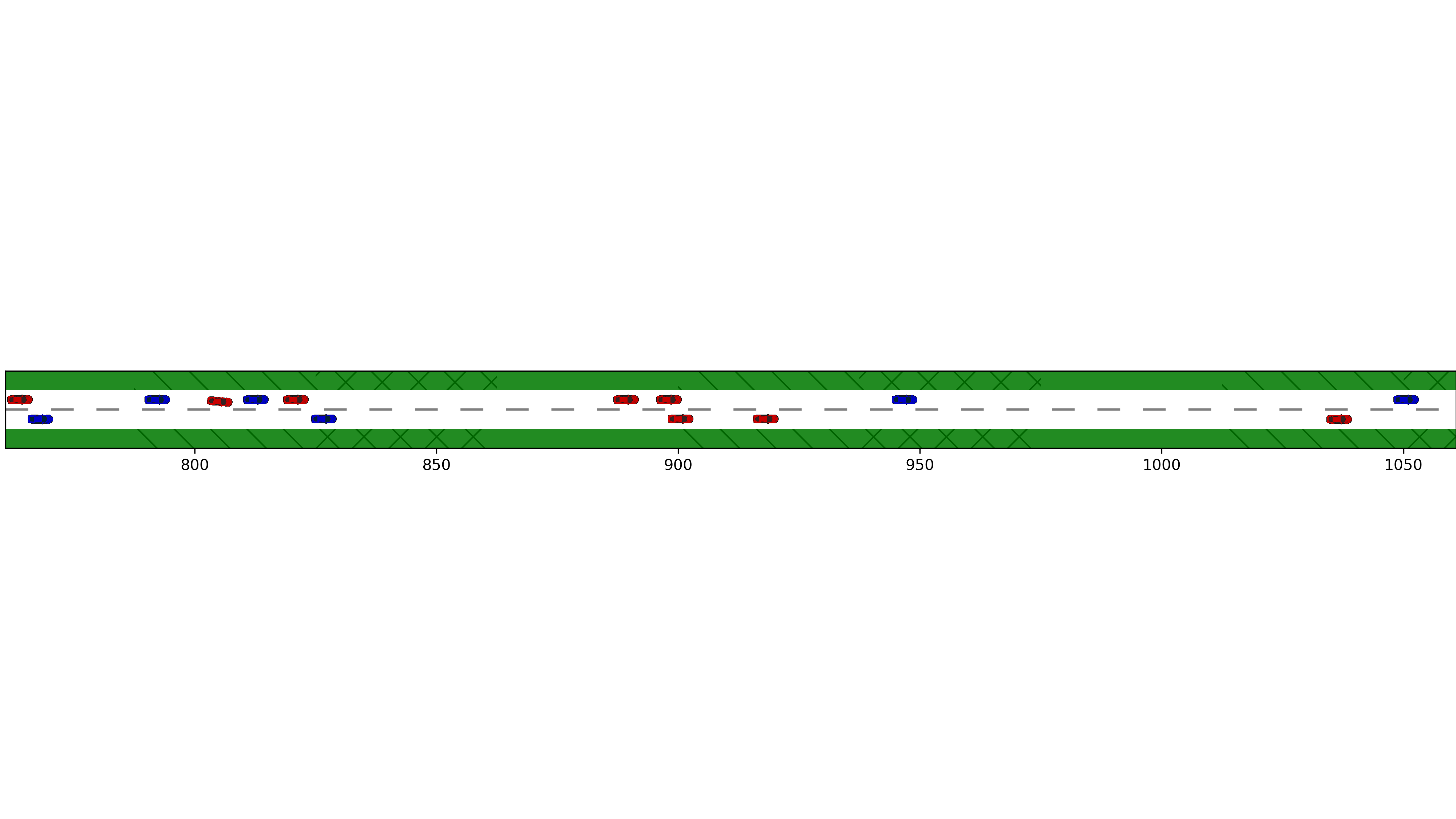}
    \includegraphics[width=0.99\columnwidth,trim={0 230px 0 245px},clip]{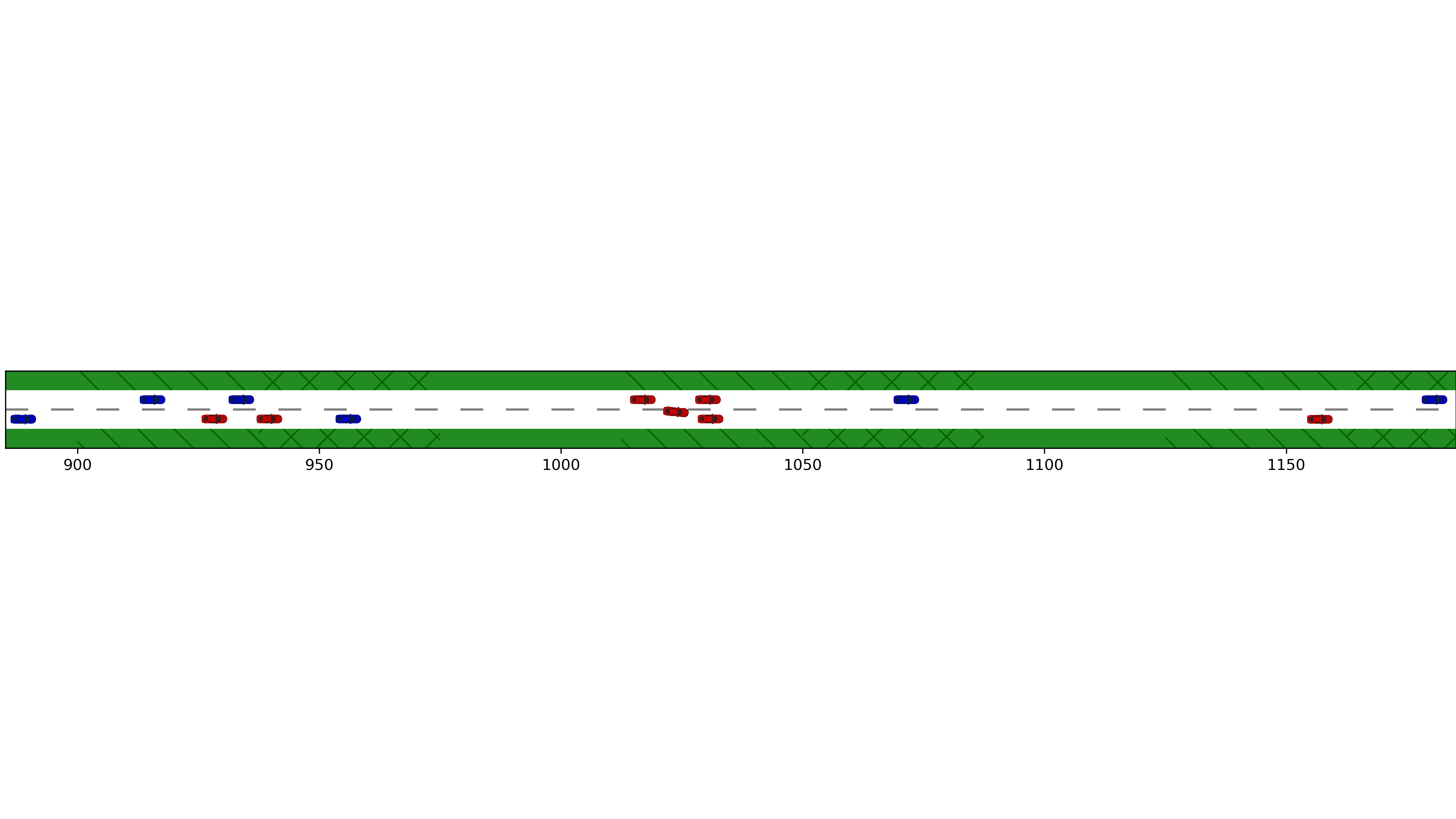}
    \includegraphics[width=0.99\columnwidth,trim={0 230px 0 245px},clip]{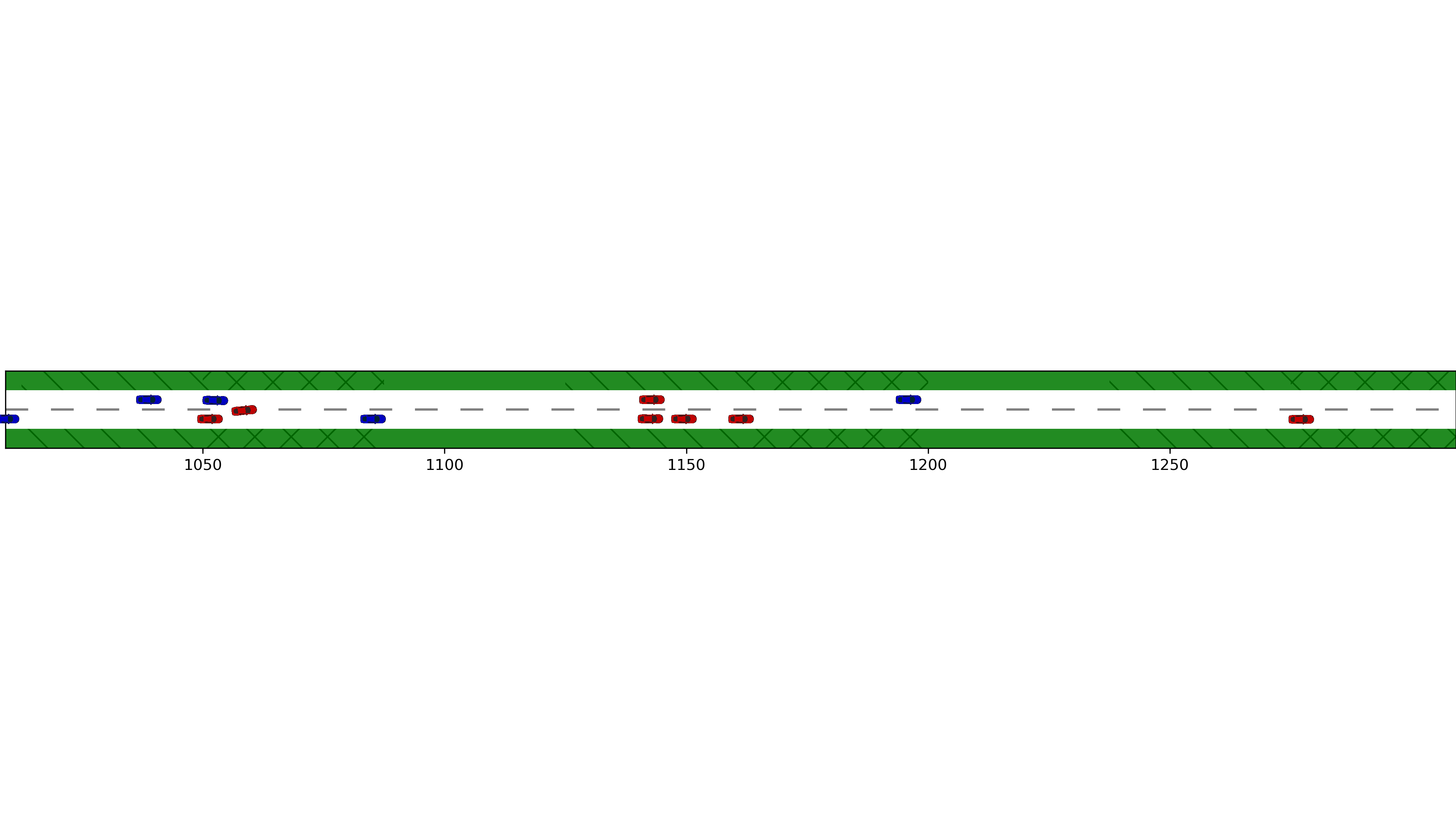}
    \includegraphics[width=0.99\columnwidth,trim={0 230px 0 245px},clip]{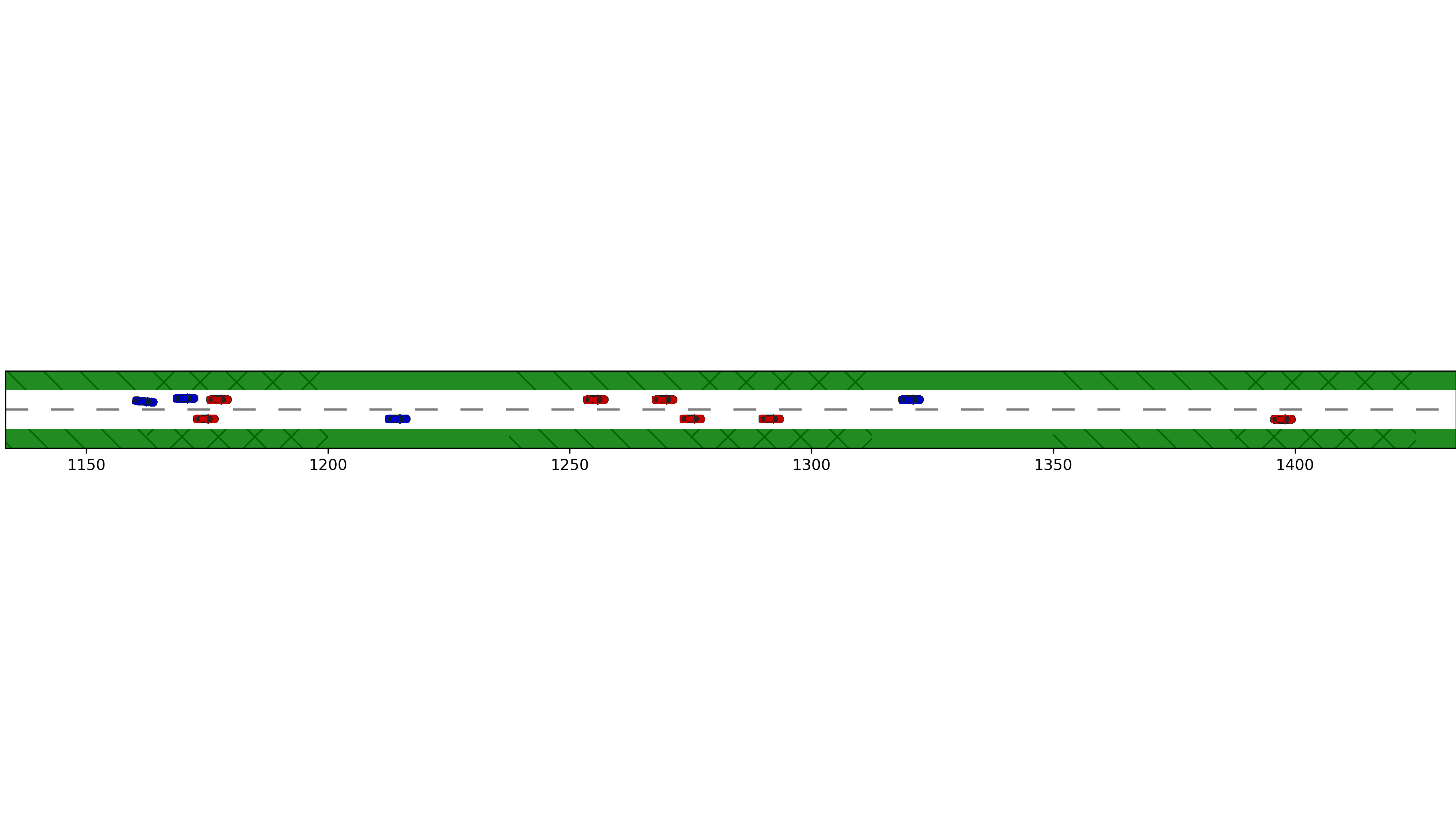}
    \includegraphics[width=0.99\columnwidth,trim={0 230px 0 245px},clip]{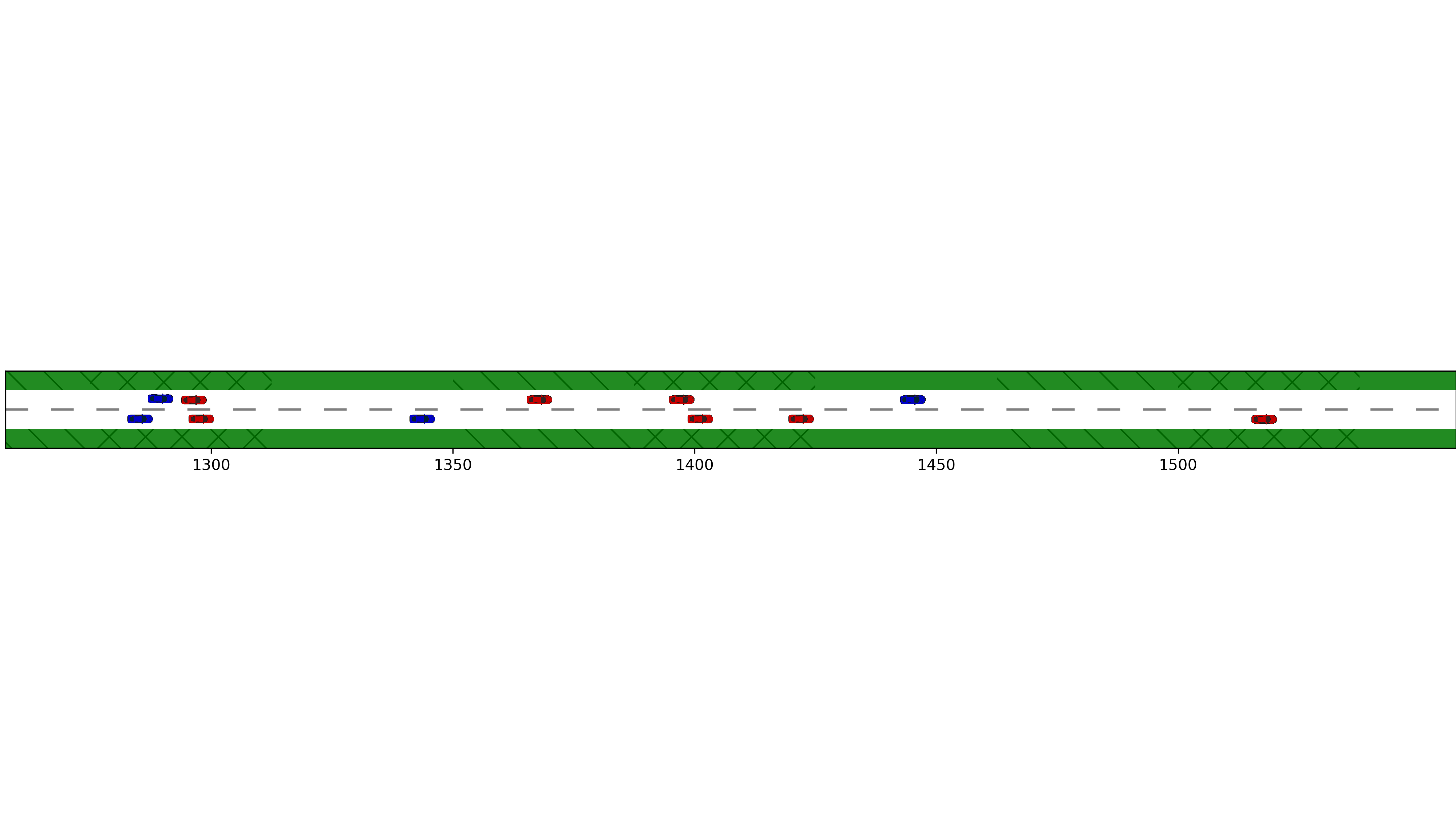}
    \caption{Simulation of 24 agents with varying SVO}
    \label{fig:iv_24_simulation}
\end{figure}

\subsection{Warm Starting and Desired Trajectories}
Given the challenging dynamics and collision avoidance constraints, we need to provide both warm starts and various possible trajectories to follow. If warm starts are not provided, our optimization solver may return with no feasible solutions.  
Likewise, if only a single desired trajectory is provided, the vehicle will penalize new maneuvers that maybe needed for passing. 
For warmstarting, we provide a pre-computed bank of initial control inputs $u_{warm}$ and states $x_{warm}$ for warmstarting the solver. 
Specifically, we provide either trajectories that are dynamically feasible from initial warm controls $u_{warm}$ (and simulate the evolution of state $x_{warm}$) or first compute geometrically feasible states 
 $x_{warm}$ and estimate corresponding (but potentially dyanmically infeasible) control inputs $u_{warm}$.

\begin{figure}[htb]
    \centering
    \includegraphics[width=0.99\columnwidth, trim={0, 0, 0, 250px}, clip]{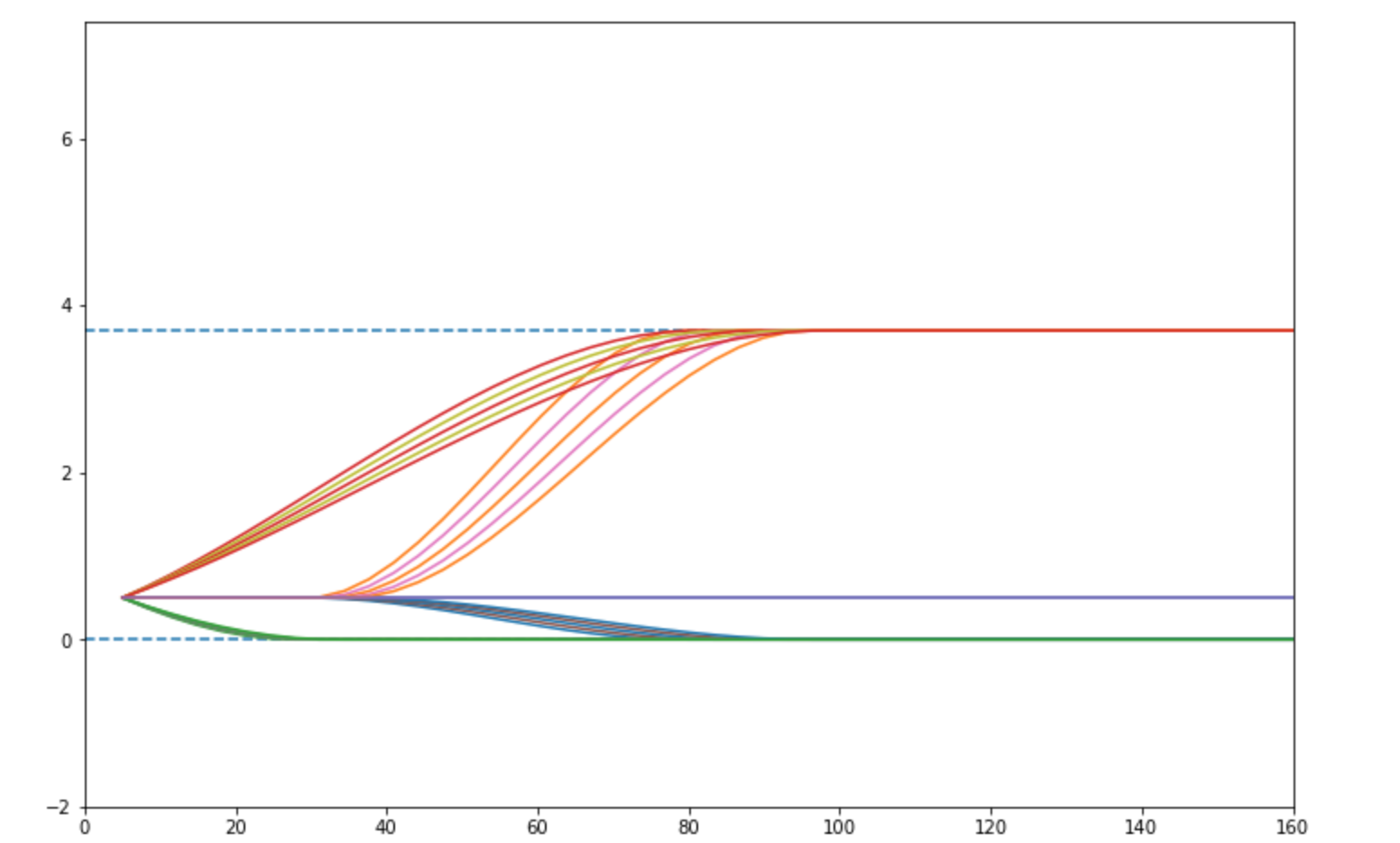}
    \caption[Desired Trajectory Polynomials]{Desired Trajectories Parameterized by Piecewise Polynomials.  Each agent considers multiple possible maneuvers to allow for contour tracking of lane following and lane changing maneuvers.}
    \label{fig:traj_desired}
\end{figure}

Each desired state trajectory is parameterized as following
\begin{align}
    x(s) &= f_1(s) + f_2(s - s_1) + f_3(s - s_1 - s_2 ) \\
    y(s) &= g_1(s) + g_2(s - s_1) + g_3(s - s_1 - s_2 ) \\ 
    \phi(s) &= h_1(s) + h_2(s - s_1) + h_3(s - s_1 - s_2 )
\end{align}
where $f, g, h$ are cubic piecewise polynomials of the form
\[ f_n = \begin{cases}
        c_3 s^3 + c_2 s^2 + c_1 s + c_0 & 0 \leq s \leq s_n \\
        0 & else
        \end{cases}
\]
and $c_0, c_1, c_2, c_3,$ are polynomial specific coefficients that are computed to fit a lane-change maneuver. 
These together with the piecewise polynomials allows for multiple smooth desired trajectories that can include both a lane following portion and a lane change portion.
Figure~\ref{fig:traj_desired} shows an example bank of desired trajectories for the ego vehicle that is considered during optimization. 
Desired trajectories are generated to consider maintaining current lane, switching lanes, and finishing mid-lane change. 

\subsection{Choosing a Feasible Solution in Finite Time}

One shortcoming of local interior point solvers such as IPOPT~\cite{wachter2006implementation} is that for nonlinear problems, the solver can take a long time to converge to a solution. 
For real-time systems, a solution should be returned within a fixed time period so that the vehicle can execute the commands. 
We fix a compute time $t_{c}$ after which a solution must be returned. We interrupt the solver at $t_{c}$ which may yield a solution $x_{c}$ which may be infeasible.  
As such, we compute the feasibility of the solution as $|g_x| = g(x_c)^T g(x_c)$ where $g(\cdot)$ consists of all constraints (dynamics, collision avoidance, control constraints) found in \eqref{eq:iv_br_constraints}. We denote all returned feasible solutions $\mathcal{X}_{f} =(x: |g_x| \leq \epsilon_f)$ and infeasible solutions $\mathcal{X}_{nc} = (x: |g_x| > \epsilon_f$ where $\epsilon_f$ is a feasibility threshold.  We then select a trajectory as follows
\begin{equation}\label{eq:feasible_solutions}
    x^* = \begin{cases} 
      \min_{x \in x_c} cost(x_c) &  |\mathcal{X}_f| > 0  \\
     \min_{x \in x_c} cost(x) + k_{slack} |g_x| & |\mathcal{X}_f|=0
   \end{cases}
\end{equation}
where $|\mathcal{X}_f$ is the number of feasible solutions. Equation~\eqref{eq:feasible_solutions} ensures that we select feasible solutions when they exist, otherwise, choosing an infeasible solution that has the smallest constraint violations.

\section{Results}
\subsection{Traffic Simulations}
\begin{figure}[htb]
        \centering
        \includegraphics[width=0.849\columnwidth]{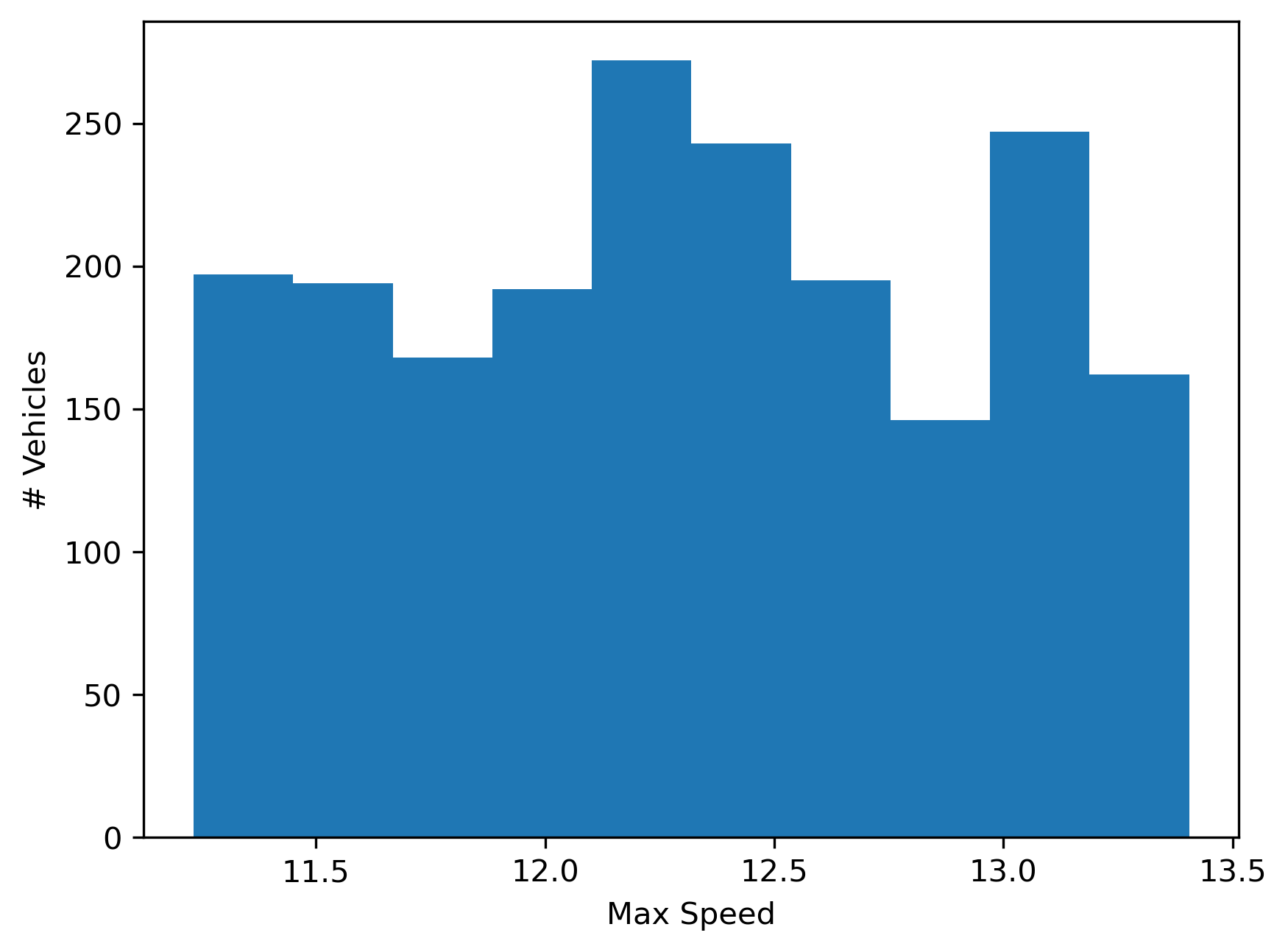}
        \caption{Vehicle Desired (Max) Speed.  Histogram of desired maximum speeds for each vehicle in system.}
        \label{fig:vehicle_speeds}\label{fig:density_speeds}
\end{figure}

Simulations of 24 vehicles running Iterative Best Response with Imagined Shared Control is repeated under various social value population, for a total of 65 separate simulations. 
The simulated environment and MPC are implemented in Python, using the CasADi framework with IPOPT~\cite{wachter2006implementation} solver, and experiments deployed on MIT's SuperCloud running on Intel Xeon
Platinum 8260.
Figure~\ref{fig:iv_24_simulation} shows a snapshot of the traffic simulation at different time steps of the simulation.
For simplicity, we assume that vehicle's pairwise SVO's are homogeneous $\theta_{ij} = \theta_i$ which is a reasonable assumption for normal highway drive, and limit the possible SVOs to the range of prosocial $\theta_i = \pi/4$ to egoistic $\theta_i \approx 0$, values typically seen in laboratory settings~\cite{garapin2015does}.  
Vehicles are initially placed in the system according to a Poisson distribution with a desired road density $\rho$, where $\rho = 3000$ cars per hour. 
{
    Vehicles are randomly assigned a desired speed ranging from 11.2 m/s to 13.4 m/s. The distribution of desired speeds for all vehicles are shown in Figure~\ref{fig:vehicle_speeds}.
}

\begin{figure}[htb]
    \centering
    \includegraphics[width=0.99\columnwidth,trim={0 1435px 0px 0px}, clip]{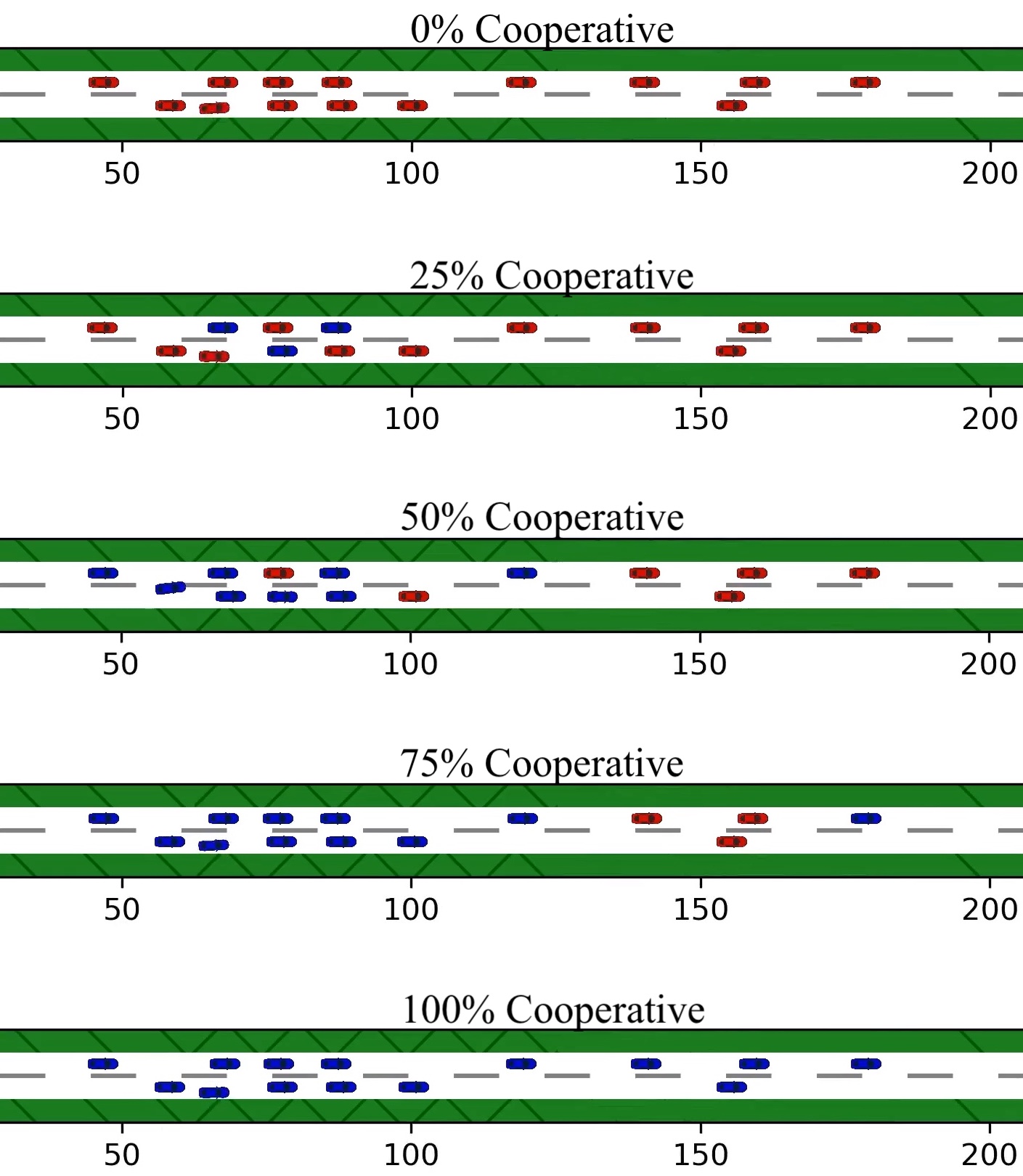}
    \includegraphics[width=0.99\columnwidth,trim={0 1090px 0px 305px}, clip]{iv2023_optimalsvo/figs/simulation/seed/seed10023_crop.jpg}
    \includegraphics[width=0.99\columnwidth,trim={0 745px 0px 650px}, clip]{iv2023_optimalsvo/figs/simulation/seed/seed10023_crop.jpg}
    \includegraphics[width=0.99\columnwidth,trim={0 405px 0px 985px}, clip]{iv2023_optimalsvo/figs/simulation/seed/seed10023_crop.jpg}
    \includegraphics[width=0.99\columnwidth,trim={0 	70px 0px 1330px}, clip]{iv2023_optimalsvo/figs/simulation/seed/seed10023_crop.jpg}           
    \caption{Varying Proportion of Cooperative Agents in Simulation. Colors correspond to prosocial (blue) or egoistic agents (red) in the simulation.}
    \label{fig:prop_cooperative}
\end{figure}

\subsection{Performance Metrics}
We are interested in studying the overall traffic flow in our system for different populations of human drivers. We measure both the individual performance and the traffic-wide performance by measuring average vehicle speed $V_i =\frac{1}{N} \sum_{t=1}^N ||v_{i,t}||$ of agent $i$ in a given experiment for $N=500$ timesteps. To closely compare between SVO populations, we repeat a given simulation (fixing initial position and desired speed) with different individual vehicle SVO settings. Then we compare the individual and population compared to the counterfactual baseline performance of a fully egoistic system ($0\%$ cooperative agents) where now our performance metrics are
\begin{align}
    \text{Individual Speed Improvement (ISI)} &= \frac{V_i}{V^e_{i}} \\ 
    \text{Population Speed Improvement (PSI)} &= \frac{\sum_i V_i}{\sum_i V^{e}_{i}}
\end{align}
where $V_i$ is the average speed of agent $i$ and $V^{e}_{i}$ is the average speed of $i$ in the baseline configuration. 

An individual speed improvement $ISI>1$ corresponds to individuals improving their travel efficiency under the current simulation settings, by achieving a higher average speed during the duration of the simulation. In contrast, a population performance $PSI>1$ corresponds to a situation where the overall flow of the entire system increased in the current simulation compared to the baseline. Typically the average individual performance and population performance will be similar but not equal depending on the distribution of improvements.

\begin{figure}[thb]
\centering
\begin{subfigure}[b]{0.89\columnwidth}
    \centering
    \includegraphics[width=0.99\columnwidth,trim={0px, 0px, 0, 0},clip]{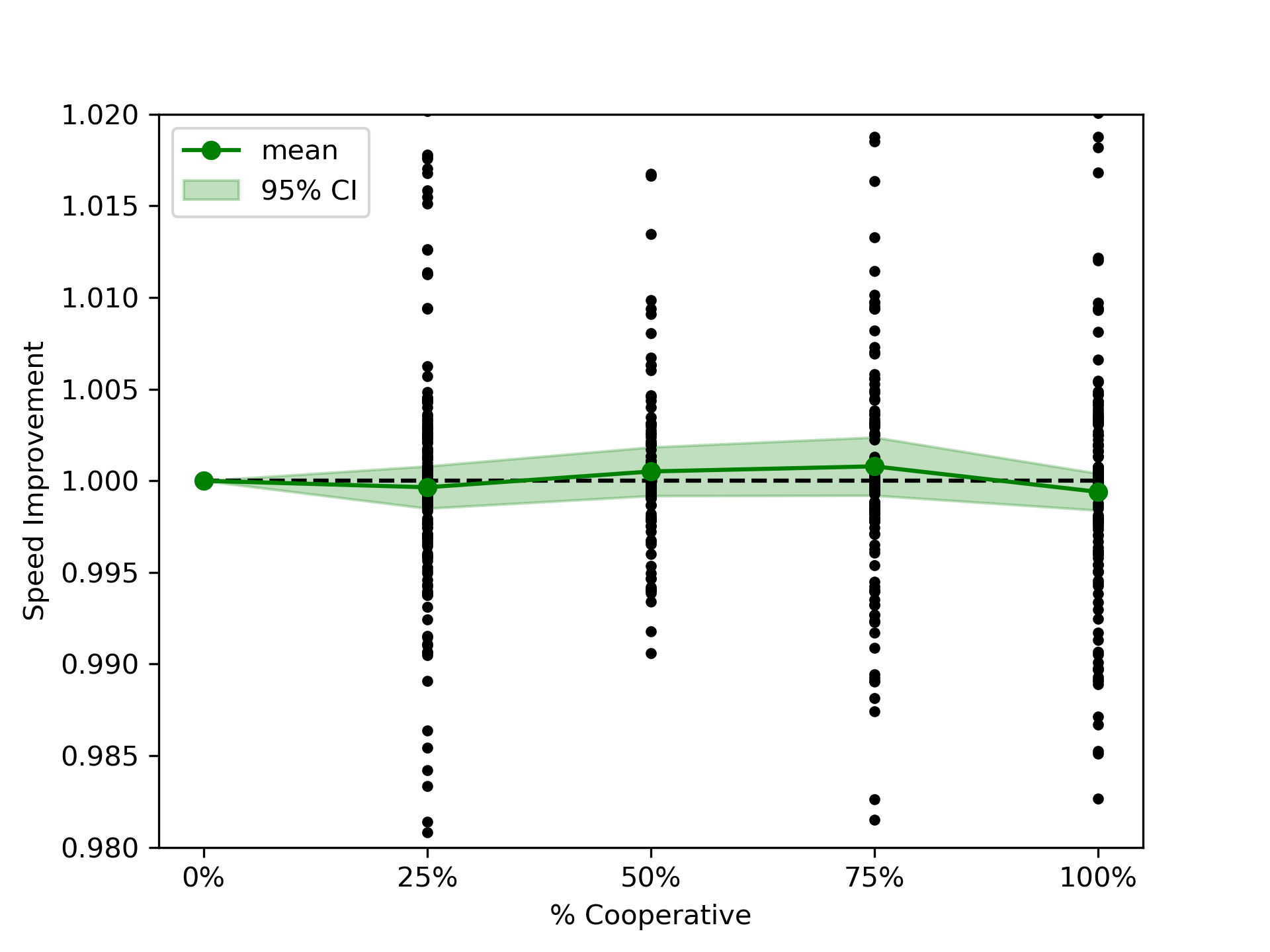}
    \caption{Individual}
    \label{fig:distance_ind}
\end{subfigure}
\begin{subfigure}[b]{0.89\columnwidth}
    \centering
    \includegraphics[width=0.99\columnwidth,trim={0px, 0px, 0, 0},clip]{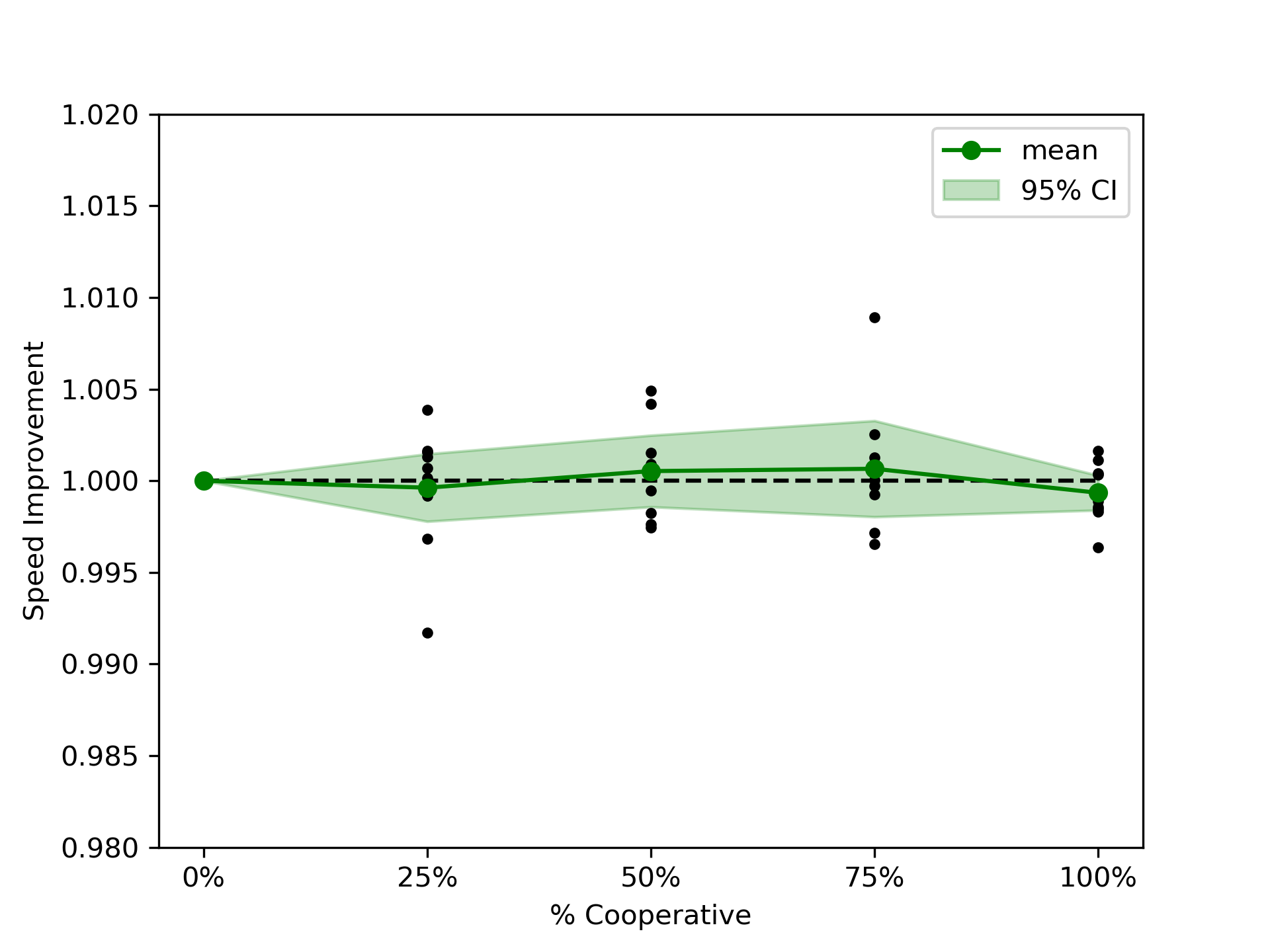}
    \caption{System}
    \label{fig:distance_pop}
\end{subfigure}
\caption[Performance of vehicles compared to baseline of no egoistic vehicles]{Performance of vehicles compared to baseline of no egoistic vehicles. \textbf{(a)}~Each datapoint is an individual driver's speed improvement. \textbf{(b)}~Each datapoint is the entire traffic system average speed improvement.}
\label{fig:distance_indiv_pop}
\end{figure}

\subsection{Varying Cooperative Agents}
We vary the proportion of agents that are cooperative ($\theta_{ij} = \pi/4)$ or egoistic ($\theta_{ij} \approx 0$) for $10$ different initial condition. 
Figure~\ref{fig:prop_cooperative} shows the same seeded simulation with five different proportions of cooperative agents: $p_{cooperative} = [0\%,25\%, 50\%, 75\%, 100\%]$ by varying a subset of agents' SVOs.
Figure~\ref{fig:distance_indiv_pop} plots the relative performance for each individual agent (Fig.~\ref{fig:distance_ind}) and the system as a whole (Fig.~\ref{fig:distance_pop}).
For low levels of cooperation, $p_{cooperative} \leq 50\%$ there appears to be a slight degradation in individual performance and a slight improvement at higher levels of cooperation., with a slight final degradation at $100\%$ cooperative agents. 
However, as a whole, the individual and system-wide performance does not improve significantly as more cooperative drivers enter the system. One potential reason for the lack of significant improvement is that the increased cooperation comes with additional inefficiencies due to lane changing and slowing down for other vehicles that reduces the overall improvement.  As a result, while some agents may improve due to cooperation, an equal number of agents may reduce the performance. For example, the reduction at 100\% cooperation may be due to not enough egoistic vehicles present to capture the benefits of the cooperative agents.

\begin{figure}[htb]
    \centering
    \includegraphics[width=0.99\columnwidth]{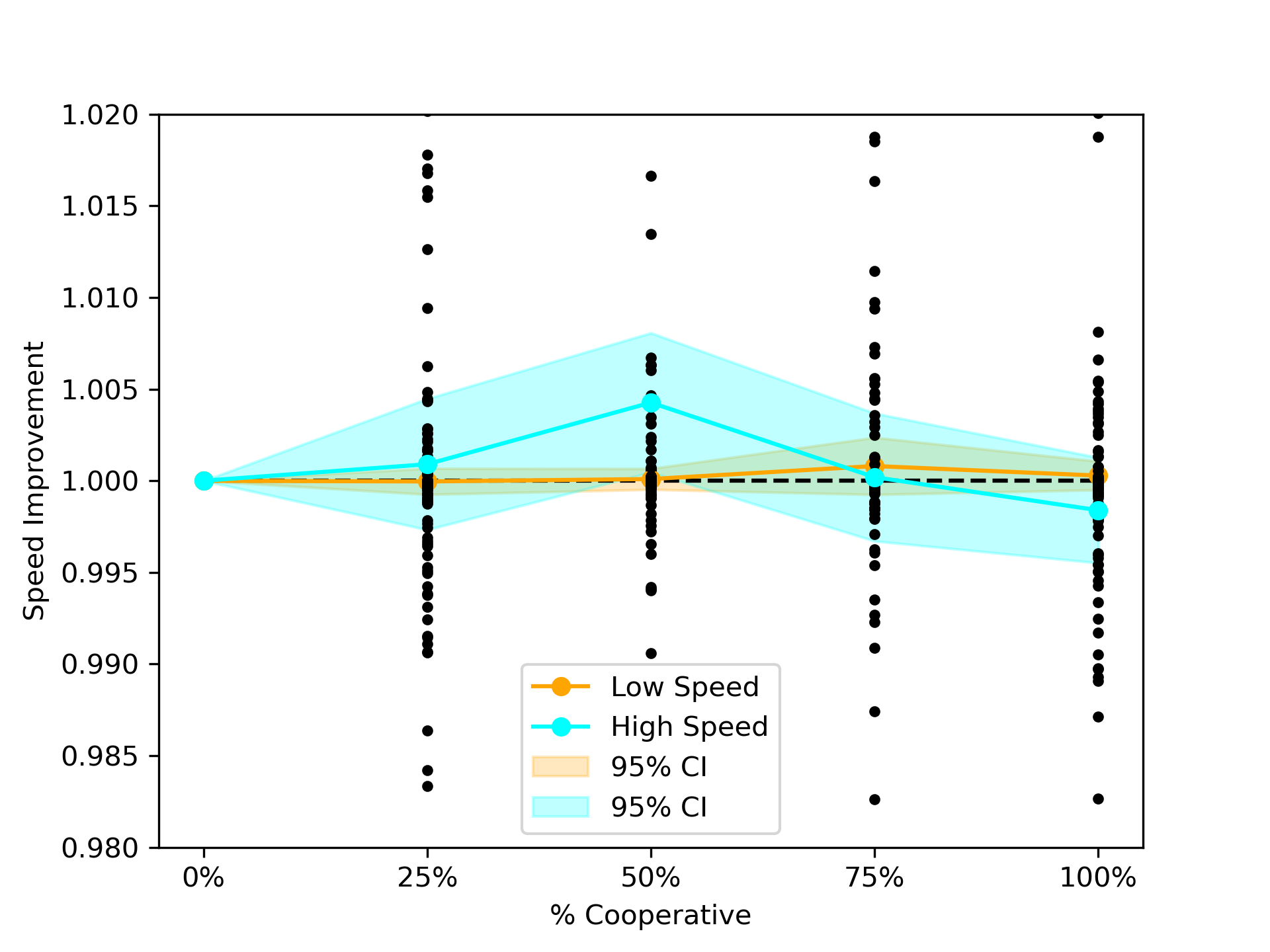}
    \caption[High Speed vs. Low Speed Drivers]{High Speed vs. Low Speed Drivers. Low speed drivers are only minimally impacted by cooperation drivers, whereas high speed drivers performance varies as cooperation levels increase.}
    \label{fig:highlowspeed}
\end{figure}

\subsection{Impact on Individual Driver Type}

To better understand the impact of cooperative agents, we explore the individual performance of the drivers to understand if different subpopulations are impacted more by more cooperative roads. 
First, in Fig.~\ref{fig:highlowspeed}, we compare drivers with a high speed limit to those with lower desired speeds to see if they have differing impacts on performance. 
As expected, low speed drivers have very little improvement on speed given that low-speed drivers are less likely to consider speeding up in cooperative scenarios. 
In contrast, high speed drivers have more variability in speed performance with an apparent speed-up when the at 50\% cooperative population of drivers.

\begin{figure}[htb]
    \centering
    \includegraphics[width=0.99\columnwidth]{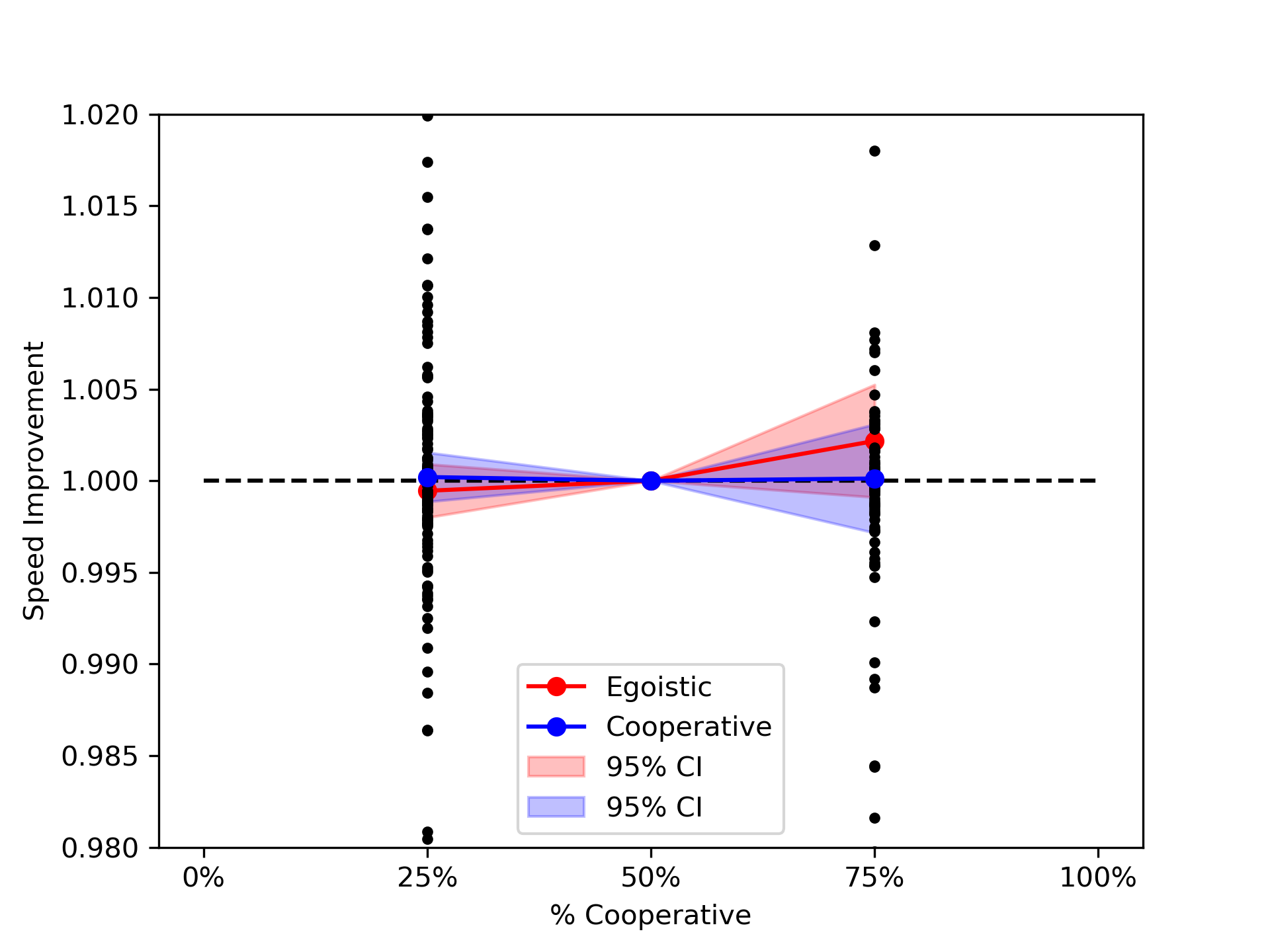}
    \caption[Effect on Prosocial and Egoistic Agents]{Effect on Prosocial and Egoistic Agents. Egoistic agents benefit from more cooperative traffic where as cooperative agents see little change in performance.}
    \label{fig:coop_noncoop}
\end{figure}

Second, we explore the impact on drivers that remain egoistic or prosocial throughout the experiments to see whether their performance is negatively or positively impact by the change in SVO of the other drivers. 
Figure~\ref{fig:coop_noncoop} compares agents who remain egoistic or prosocial during the experiments and compare them against the baseline of the population when agents are 50\% prosocial. 
Note that simulations with 100\% or 0\% cooperative agents are excluded since those simulations do not contain egoistic or prosocial agents, respectively.
We can see that prosocial agents do not get impacted by the change in population cooperation, however, egoistic agents do improve performance as the cooperation levels increase. 
This highlights that egoistic agents gain benefits when an increasing proportion of agents in the system are prosocial.

\begin{figure}[htb]
\centering
\begin{subfigure}[b]{0.99\columnwidth}
    \centering
    \includegraphics[width=0.99\columnwidth,trim={0px, 0px, 0, 0},clip]{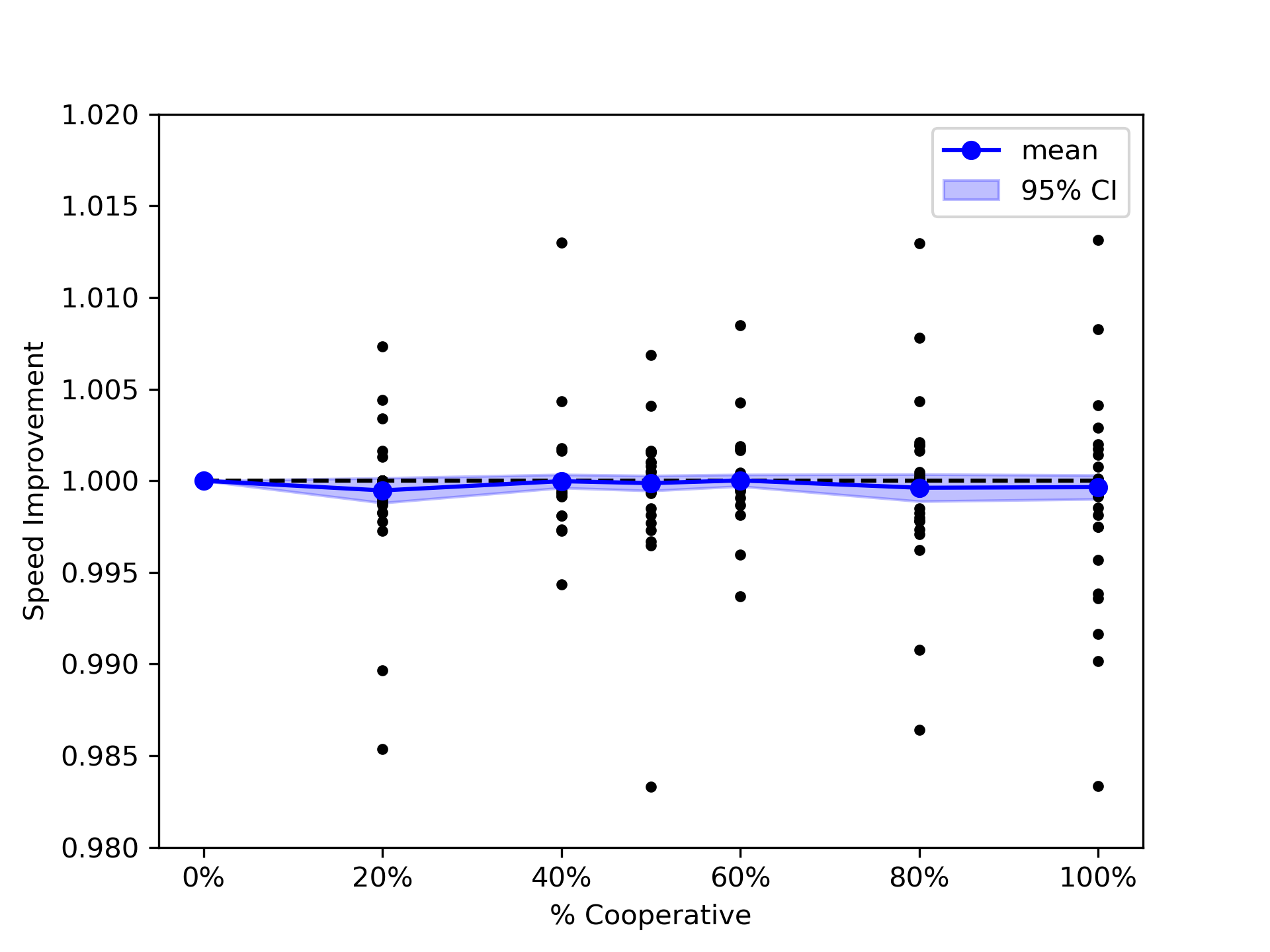}
\end{subfigure}
\caption[Increasing Shared Control]{Increasing Shared Control. Experiments are repeated with a more cooperative version of IBR by increasing the number of vehicles considered in shared control. The increased consideration does not improve the individual performance in the highway setting.}
\label{fig:control_distance_nc2}
\end{figure}

\subsection{Varying Shared Control and Traffic Density}
To better understand the joint impact of algorithm parameters and vehicle SVO, we repeat the experiments on a subset of the population (eight vehicles). 
First, varying the size of shared control has the potential to increase the problem complexity, leading to slower solve time and possible instability, while also converging to more cooperative solutions for each driver. 
We increase the shared control from $n_{sc}=1$ to $n_{sc}=2$ and re-run the experiments with different SVO populations. 
Fig.~\ref{fig:control_distance_nc2} shows the results for the the increased neighborhood of shared control.
We see that there is a slight reduction in performance with additional cooperation, possibly due to the instability at higher number of agents co-planning together.

\begin{figure}[htb]
\centering
\begin{subfigure}[b]{0.99\columnwidth}
    \centering
    \includegraphics[width=0.99\columnwidth,trim={0px, 0px, 0, 0},clip]{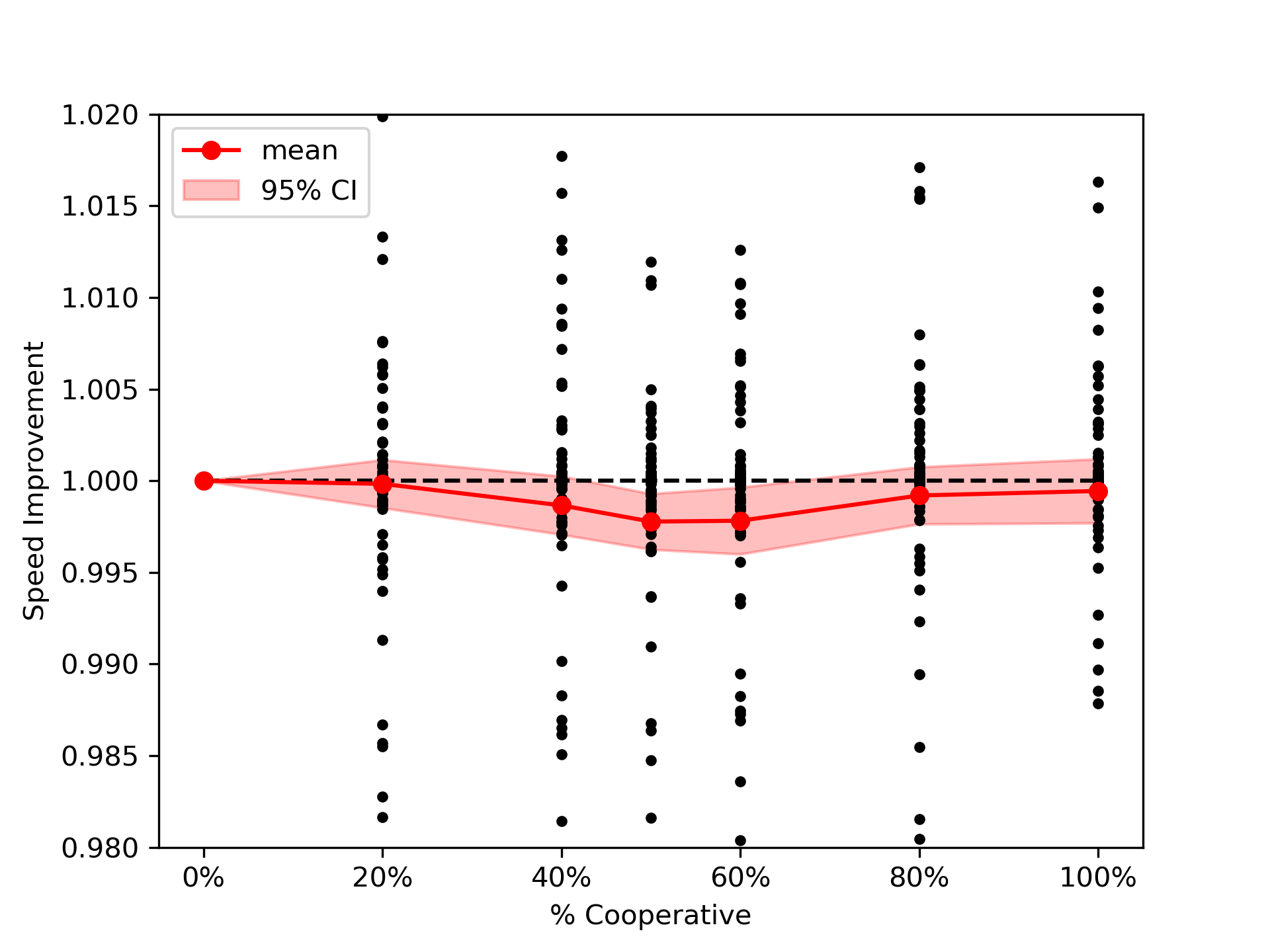}

\end{subfigure}

\caption[High Density Individual Performance]{High Density Individual Performance. Experiments are repeated with a higher density traffic setting. Performance degrades with additional cooperation due to saturated roads that prevent effective cooperation.}
\label{fig:control_distance_highdensity}
\end{figure}

Increasing density has the potential to increase the benefit of cooperation, due to the need to coordinate to drive through traffic, or the potential to reduce the benefit given the highly constrained driving scenario. 
Figure~\ref{fig:control_distance_highdensity} shows results for twice the density as shown previously. 
At higher traffic density, we see a consistent reduction in performance due to cooperation, peeking at $50\%$ of the population cooperative. A possible explanation for this observation is that at high densities, the vehicles do not have room to take advantage of prosocial agents. Instead, forming a traffic bottleneck and incurring the extra effort to cooperate.

\section{\Conclusion{}}
We evaluate the system wide performance for a multiagent semi-cooperative planning for human and autonomous vehicles on the road. 
We deploy a low-level model predictive controller that conducts implicit teaming with neighboring vehicles to achieve cooperative maneuvers. 
Simulations show that while individual performance improves slightly as more agents become cooperative, the overall effect is not statistically significant. In addition, we consider the performance of individual sub-populations of drivers and see that egoistic agents do benefit from prosocial drivers.   
This \work{} highlights the need for additional strategies and research in implicit coordination for autonomous vehicles and the limitations of semi-cooperative behaviors when considering low-level control.

\addtolength{\textheight}{-2in}   

\bibliographystyle{IEEEtran}
\bibliography{IEEEabrv,references}

\end{document}